\documentclass[10pt,twocolumn,letterpaper]{article}

\usepackage[pagenumbers]{cvpr} % To force page numbers, e.g. for an arXiv version
\usepackage[T1]{fontenc}
\AtBeginDocument{\hfuzz=0.1pt\vfuzz=0.1pt}
\let\cvpreqref\eqref
\usepackage{amsmath,amsfonts,bm}

\def\eqref#1{equation~\ref{#1}}
\def\1{\bm{1}}

\DeclareMathAlphabet{\mathsfit}{\encodingdefault}{\sfdefault}{m}{sl}
\SetMathAlphabet{\mathsfit}{bold}{\encodingdefault}{\sfdefault}{bx}{n}

\let\eqref\cvpreqref
\usepackage{enumitem,multirow,amsthm,bbm,colortbl,arydshln,tikz}
\usepackage[most]{tcolorbox}

\newcommand{\selfreactivity}{self-reactivity}
\newcommand{\reactdrive}{ReactDrive}

\definecolor{boschblue}{RGB}{0,79,112}
\definecolor{boschred}{RGB}{237,0,7}

\newcommand{\myparagraph}[1]{\vspace{0pt}\noindent{\bf{#1}}~~}
\definecolor{resultBestBg}{HTML}{E9F3FF}
\definecolor{resultSecondBg}{HTML}{FFF5E6}
\newcommand{\rankfirst}[1]{\cellcolor{resultBestBg}\textbf{#1}}
\newcommand{\ranksecond}[1]{\cellcolor{resultSecondBg}\underline{#1}}
\theoremstyle{definition}

\tcbuselibrary{skins}
\newtcolorbox{takeaway}{
  enhanced, breakable,
  colback=gray!8, colframe=gray!8,
  coltitle=black,
  boxrule=0pt, arc=2pt,
  left=5pt, right=5pt, top=3pt, bottom=3pt,
  before skip=4pt, after skip=4pt,
  fonttitle=\bfseries\sffamily,
  attach title to upper={},
  title={Takeaway:~~},
}
\definecolor{cvprblue}{rgb}{0.21,0.49,0.74}
\usepackage[pagebackref,breaklinks,colorlinks,allcolors=cvprblue,hypertexnames=false]{hyperref}
\def\confName{CVPR}
\def\confYear{2027}
\title{ReactDrive: Self-Reactive Driving World Model}
\author{
Yifan Wang$^{1}$%
\hyperlink{author-notes}{\textcolor{black}{$^{\dagger}$}}%
{%
\thanks{\protect\hypertarget{author-notes}{}This work was done during an internship at Bosch.\quad $^{\ddagger}$Project lead.}} \quad
Xiaoqi Wang$^{2,3}$\hyperlink{author-notes}{\textcolor{black}{$^{\ddagger}$}}\quad
Cho-Ying Wu$^{2,3}$ \\
Wenbin He$^{2,3}$ \quad
Chenyu You$^{1}$ \quad
Liu Ren$^{2,3}$ \\[6pt]
{
$^{1}$Stony Brook University \quad
$^{2}$Bosch Research North America} \\
{
$^{3}$Bosch Center for Artificial Intelligence (BCAI)}
}

\begin{document}
\setcounter{footnote}{1}
\maketitle
\begin{abstract}
The world evolves through the dynamics of its constituent entities and their interactions, with changes in one entity influencing the behavior of others.
To simulate this evolution under different actions, interactive world models need to capture the resulting agent responses as part of the world's own dynamics.
In autonomous driving, errors in modeling these responses can misrepresent maneuver safety and undermine the reliability of generated videos used for policy evaluation and learning.
To formalize this intrinsic requirement for driving world model, we define \textbf{\selfreactivity{}} as the capability to model agent responses from observed history and a specified ego action, and generate visual futures that reflect those responses without prescribed future agent trajectories.
To realize this capability, we introduce \textbf{\reactdrive{}}, a driving world model that jointly learns agent dynamics and visual generation through a shared agent state representation.
Within this framework, dynamics supervision grounds a reactive dynamics prior in observed agent states, while the video objective jointly optimizes the \selfreactivity{} and visual realism.
Experiments show improved agent responses, ego controllability, and video quality, a $96.0\%$ preference for agent reactivity in human comparisons across five baselines, and better downstream planning through augmentation with generated videos.
We hope this work encourages the community to recognize \selfreactivity{} as a foundational capability for developing and evaluating interactive world models. Project page: \url{https://reactdrive.github.io/}.
\end{abstract}

\section{Introduction}

\begin{center}
\small
\textit{``All external action in the world is interaction.''}\\
\makebox[\linewidth][r]{Immanuel Kant, 1786}
\end{center}
The world evolves through the dynamics of its constituent entities and their interactions, where changes in one entity influence the behavior and evolution of others~\citep{battaglia2016interaction}. This interactive nature of world dynamics, as emphasized by Kant's observation in 1786~\citep{kant2004metaphysical}, has a fundamental implication for world modeling: simulating the consequences of an action requires accounting for how it affects other entities. World models aim to capture world dynamics to simulate possible futures under different actions~\citep{ha2018world,hafner2019learning,pmlr-v267-zhou25t}. However, faithfully modeling such dynamics is challenging, particularly in autonomous driving, where the evolution of traffic scenes depends not only on the ego vehicle's actions but also on how surrounding agents respond to them. Recent driving world models (DWMs) have made substantial progress in visual simulation, generating realistic, temporally coherent future driving scenes~\citep{Zhang_2025_ICCV,agarwal2026cosmos} that provide experience for driving policy evaluation and learning~\citep{yang2025resim,wang2026behaviorworldgen}.
However, visual realism alone does not establish whether these simulated futures faithfully capture the agent dynamics and interactions among agents~\citep{Motamed_2026_WACV}.

\begin{figure*}[t]
  \centering
 \includegraphics[width=.85\linewidth]
  {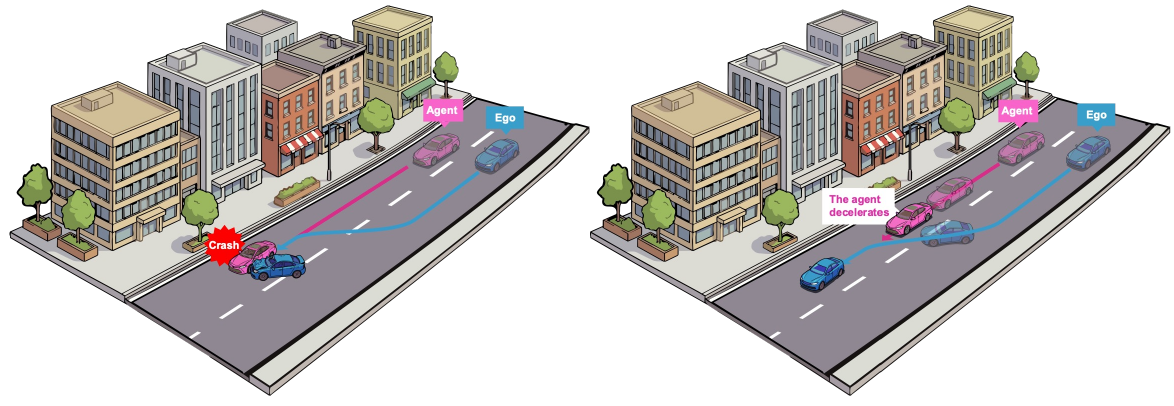}
  \vspace{-5pt}
  % \caption{TEASER}
  \caption{Agent responses are integral to world dynamics, making self-reactivity fundamental to interactive world modeling.
The illustration shows how different agent responses affect the safety of the same ego action "shift right".
Left: the agent maintains its speed, resulting in a collision.
Right: the agent decelerates, creating space for the ego vehicle to change the lane.}
  \label{fig:teaser}
  \vspace{-10pt}
\end{figure*}

In autonomous driving, how surrounding agents respond to the ego vehicle's actions can directly determine the safety and feasibility of its maneuvers~\citep{tolstaya2021identifying,Sadigh-RSS-16}.
For instance, a lane change may be safe when a neighboring vehicle yields, but dangerous when it maintains its speed or accelerates (see Figure~\ref{fig:teaser}). Failing to predict these responses can misrepresent the consequences of ego actions, undermining the reliability of simulated experience for policy evaluation and learning. Faithful simulation therefore requires DWMs to model how surrounding agents respond to ego actions, rather than merely generating visually plausible futures. We refer to this capability as \emph{\selfreactivity{}}: the ability of a DWM to model agent responses from observed scene and a queried ego action, and to generate a visual future consistent with those responses without prescribed future agent trajectories. 
Empowered by \selfreactivity{}, DWMs can move beyond realistic video generation toward interactive world simulators that faithfully reflect the consequences of the ego vehicle's actions.

Despite substantial progress in controllable video generation, existing DWMs do not necessarily possess the \selfreactivity{} required for interactive world simulation. One line of research conditions video generation on ego actions while implicitly learning agent responses through visual prediction objectives~\citep{hu2023gaia1,gao2024vista}. However, without explicit supervision of agent states, their objectives do not directly enforce whether the generated agent responses are consistent with the queried ego action. Another line of research enables fine-grained control by conditioning video generation on prescribed future 3D object layouts and agent trajectories~\citep{Gao_2025_ICCV,zhao2025drivedreamer,zhu2026other,zhou2026toward}. While effective for controllable generation, this paradigm requires plausible future agent trajectories to be annotated for each scenario, introducing substantial annotation cost when simulating diverse driving scenarios at scale.
Recent driving simulators have further demonstrated the importance of modeling agent responses for policy learning~\citep{yan2026causaldrive,wang2026behaviorworldgen}, highlighting the need to integrate this capability into physically grounded and visually coherent world generation. Building on these advances, we pose a central challenge: \emph{How can visual world models infer physically grounded agent responses and generate future scenes consistent with these responses?}

To address this challenge, we introduce \reactdrive{}, a driving world model to simulate the driving scene consistent with the interactive world dynamics. Our key idea is to enable the world model itself to learn interactive dynamics, so that it can infer how surrounding agents react to ego actions and generate visual futures consistent with those responses. Specifically, we develop a reactive dynamics prior to infer future agent states from the observed scene history and a queried ego action. During training, ground-truth agent trajectories under factual ego actions provide direct supervision for this prior, grounding the inferred responses in observed agent dynamics rather than relying on visual supervision alone. These inferred responses further guide video generation, enabling the model to produce realistic future scenes that reflect the underlying agent dynamics. Through joint dynamics and video supervision, \reactdrive{} learns to connect agent response inference with visual generation. At inference, \reactdrive{} generates future driving scenes that reflect surrounding-agent responses to different ego actions, without requiring predefined future agent trajectories.

We comprehensively evaluate \reactdrive{} on the Waymo Open Dataset~\citep{sun2020scalability} and nuPlan~\citep{caesar2021nuplan} in terms of ego controllability, agent reactivity, visual quality, and safety. \reactdrive{} achieves the lowest agent average displacement error (ADE), reducing by 9.6\% on Waymo and 25.5\% on nuPlan relative to Cosmos3~\citep{agarwal2026cosmos}.
These quantitative improvements are further supported by a human evaluation, in which participants prefer \reactdrive{} for agent reactivity in $96.0\%$ of $623$ pairwise comparisons across five baselines. 
To assess the benefits of generated videos for downstream planning, we train a driving planner on real data augmented with videos generated by \reactdrive{},, achieving a Predictive Driver Model Score (PDMS) of $85.1$ on NAVSIM~\citep{dauner2024navsim}, compared with $83.8$ when trained on Cosmos3 generated videos. Our major contributions are threefold:
\begin{itemize}[leftmargin=*,nosep]
    \item We establish \selfreactivity{} as an intrinsic requirement of interactive world modeling and formulate it as an explicit objective for model development and evaluation.
    \item We introduce \reactdrive{}, a self-reactive driving world model that jointly learns physically grounded inter-object dynamics and generates visually consistent future scenes.
    \item We show that \reactdrive{} produces driving scenes with more plausible agent responses and enhances downstream planning performance through our generated videos.
\end{itemize}

\section{Related Work}

\myparagraph{Controllable driving scene generation.}Driving world models generate future observations from visual history and specified ego actions~\citep{kim2021drivegan,hu2023gaia1,gao2024vista,Zhang_2025_ICCV,russell2025gaia,agarwal2026cosmos}. Additional visual and language conditions extend control to object dynamics and agent behavior~\citep{hassan2025gem,yan2026causaldrive}. However, without explicit supervision of agent states, visual objectives alone do not directly enforce agent responses consistent with the queried ego action. 
Another line introduces explicit scene representations to control road geometry~\citep{yang2023bevcontrol,swerdlow2024BEVGen} and object layouts~\citep{gao2023magicdrive,liu2024mvpbev,wen2024panacea,gao2024magicdrive3d,Gao_2025_ICCV}, with volumetric and occupancy representations providing further geometric guidance~\citep{lu2024wovogen,li2025uniscene,lu2025infinicube}. Scene reconstruction supports object and trajectory editing~\citep{ost2021nsg,yang2023unisim,wei2024editable,chen2024omnire,wang2026horizonforge}, while neural rendering and generative view synthesis produce observations along alternative camera trajectories~\citep{tonderski2024neurad,wang2024freevs,ni2025recondreamer,yan2025streetcrafter}. Beyond scene geometry, future agent trajectories can be supplied directly or generated from instructions~\citep{zhu2026other,zhao2025drivedreamer}, with further refinement before visual synthesis~\citep{park2025simsplat,zhou2026toward}. Prescribing these trajectories provides direct control over agent behavior but introduces additional construction costs when scaling simulation across diverse scenes and ego actions.

\begin{figure*}[t]
\centering
\includegraphics[width=.9\textwidth]{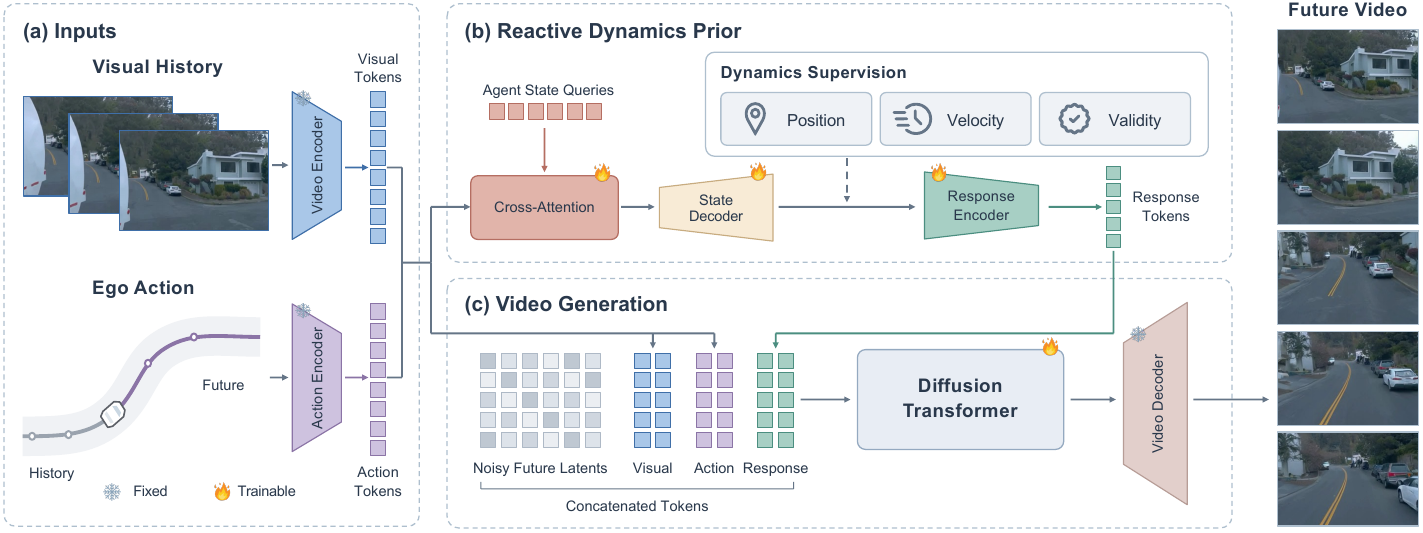}
\caption{\reactdrive{} integrates agent response modeling and visual generation within a unified world model.
(a): The encoders map visual history and ego actions to shared tokens.
(b): The reactive dynamics prior estimates the agent states and encodes them as response tokens.
(c): The diffusion transformer uses the visual, action and response tokens to generate future video.}
\vspace{-10pt}
\label{fig:pipeline}
\end{figure*}

\myparagraph{Agent dynamics modeling.}Conditional trajectory models predict how surrounding agents respond to candidate ego plans~\citep{tolstaya2021identifying,ngiam2021scene}. To capture dependencies among agents, joint prediction and interactive planning model their future behaviors together~\citep{sun2022m2i,jiang2023motiondiffuser,huang2023gameformer}. Extending from prediction to simulation, generative agent models produce evolving scenes in physical state space~\citep{zhang2023trafficbots,zhou2024behaviorgpt,wu2024smart,tan2025scenediffuser,chang2025langtraj}. To provide visual experience for policy evaluation and learning, simulation frameworks combine agent dynamics with sensor generation~\citep{yang2024drivearena,tian2026simscale,wang2026behaviorworldgen}. A complementary direction uses frozen visual world representations to model reciprocal agent responses for planning~\citep{ma2026rewam}. 
However, plausible agent trajectories do not ensure that generated videos faithfully reflect the corresponding responses. This motivates treating agent dynamics and visual generation as a unified learning problem, aligning modeled responses with their visual expression in generated futures.

\section{Method}
\label{sec:methods}
Interactive world simulation requires modeling world dynamics under ego actions and generating visual futures that reflect the responses of surrounding agents.
To this end, we propose \reactdrive{}, a self-reactive driving world model that simulate the driving scene with plausible and reactive agent responses to ego actions.
We first formulate this unified framework of reactive driving world model in Section~\ref{sec:preliminaries}.
Within this framework, we develop a reactive dynamics prior (Figure~\ref{fig:pipeline}b) that models agent responses through agent states in Section~\ref{sec:reactdrive}.
Then, this agent response guides the video generator (Figure~\ref{fig:pipeline}c) to produce more physically-grounded video in Section~\ref{sec:visual_generation}. Finally, Section~\ref{sec:reactdrive_learning} describes joint training with dynamics supervision and the video objective, followed by inference for simulating future scenes under specified ego actions.

\subsection{Problem Formulation}
\label{sec:preliminaries}

Self-reactivity is a fundamental capability for interactive world simulation. However, enabling world models to learn the interactive dynamics in the world and generate consistent visual futures following this dynamics remains a significant challenge. 
Given a queried ego action, the world model must infer how surrounding agents respond, even though their future behaviors are not specified~\citep{tolstaya2021identifying}. 
Moreover, visual observations reflect changes in both agent states and the ego vehicle's position and orientation~\citep{yang2023emernerf}, making the underlying agent dynamics difficult to isolate through video supervision alone.
To address these challenges, \reactdrive{} introduces a reactive dynamics prior that explicitly models agent responses through agent states. This enables direct supervision using reference agent trajectories during training and provides physically grounded response tokens to guide visual generation.

Let $\mathbf{X}=(\mathbf{x}_{-H+1},\ldots,\mathbf{x}_{0})$ denote observed frames and $\mathbf{Y}=(\mathbf{x}_{1},\ldots,\mathbf{x}_{T})$ the future observations.
The ego action sequence $\mathbf{A}=(\mathbf{a}_{-H+2},\ldots,\mathbf{a}_{T})$ represents the observed ego trajectory followed by the queried future trajectory.
The resulting world modeling task is to represent $p(\mathbf{Y}\mid\mathbf{X},\mathbf{A})$.
As shown in Figure~\ref{fig:pipeline}a, visual history and ego actions are encoded into shared token streams for agent response modeling and video generation.
Both agent response modeling and video generation are conditioned on $\mathbf{h}=(\mathbf{z}_{\mathrm{obs}},\mathbf{A})$, where $\mathbf{z}_{\mathrm{obs}}$ denotes the observed video latents.
Introducing the future agent state sequence $\mathbf{S}$, we express the conditional distribution of future observations as
\begin{equation}
p_{\theta,\psi,\phi}(\mathbf{Y}\mid\mathbf{h})=\int p_{\theta}(\mathbf{Y}\mid\mathbf{h},G_{\psi}(\mathbf{S}))\,q_{\phi}(d\mathbf{S}\mid\mathbf{h}).
\label{eq:internal_prior_factorization}
\end{equation}
Here, the dynamics prior $q_{\phi}$ models future agent states conditioned on $\mathbf{h}$, as shown in Figure~\ref{fig:pipeline}b.
The response encoder $G_{\psi}$ then maps these states into response tokens that condition the visual model $p_{\theta}$ in Figure~\ref{fig:pipeline}c.
The video generator also retains direct access to $\mathbf{h}$, preserving scene appearance and ego trajectory information beyond the compact agent state representation.

\reactdrive{} computes future agent states through the dynamics module $F_{\phi}$ and uses the resulting estimate to define the dynamics prior:
\begin{equation}
\widehat{\mathbf{S}}=F_{\phi}(\mathbf{h}), \qquad q_{\phi}(d\mathbf{S}\mid\mathbf{h})=\delta_{\widehat{\mathbf{S}}}(d\mathbf{S}).
\label{eq:reactdrive_predictor}
\end{equation}
The point mass $\delta_{\widehat{\mathbf{S}}}$ specifies a deterministic state estimate for a fixed history and ego query, while visual generation remains stochastic.
Thus, \reactdrive{} infer its response tokens from observed history and the ego query, without prescribed future agent trajectories.

\subsection{Reactive Dynamics Prior}
\label{sec:reactdrive}
\label{sec:self_reactivity}

The reactive dynamics prior represents agent responses via agent states, connecting agent dynamics with video generation.
We first define the state representation, then compute future state sequences from the shared visual and action tokens, and finally supervise these states using reference agent trajectories.
We formulate agent positions and velocities on the ground plane in a coordinate system fixed at the last observed ego pose, providing a common spatial reference throughout the rollout.
Within this frame, each agent state comprises position, velocity, and a validity indicator:
\begin{equation}
\mathbf{S}=\{(\mathbf{s}_{t,n},m_{t,n})\}_{t,n}, \qquad \mathbf{s}_{t,n}=[\mathbf{p}_{t,n};\mathbf{v}_{t,n}]\in\mathbb{R}^{4}.
\label{eq:agent_state}
\end{equation}
Here, $\mathbf{p}_{t,n},\mathbf{v}_{t,n}\in\mathbb{R}^{2}$ denote position and velocity, respectively.
The index $t$ denotes a future frame, while $n$ identifies the same agent across future frames.
The reference label $m_{t,n}\in\{0,1\}$ indicates state validity, while the model estimates a validity probability $\widehat{m}_{t,n}\in[0,1]$.

Given this state representation, the dynamics module $F_{\phi}$ in Figure~\ref{fig:pipeline}b computes future agent state sequences from the shared visual and action tokens.
Spatial and temporal position information preserves the organization of visual observations and the order of ego actions.
Learned agent state queries attend to these tokens. The state decoder maps each updated query to a sequence of positions, velocities, and validity probabilities for one agent over the full rollout.
Conditioning on the full ego action sequence allows the agent state estimates to depend on the proposed maneuver throughout the rollout.
During training, dynamics supervision grounds the reactive dynamics prior in reference agent positions, velocities, and state validity.
For a continuous agent trajectory indexed by time $\xi$, position and velocity satisfy $\dot{\mathbf{p}}_{n}(\xi)=\mathbf{v}_{n}(\xi)$, giving
\begin{equation}
\mathbf{p}_{n}(\xi+\Delta t)-\mathbf{p}_{n}(\xi)=\int_{\xi}^{\xi+\Delta t}\mathbf{v}_{n}(s)\,ds.
\label{eq:physical_kinematics}
\end{equation}
This relation motivates supervising positions and velocities together, so that reference trajectories ground both the spatial evolution of each agent and its change.
The resulting dynamics supervision combines four complementary terms:
\begin{equation}
\ell_{\mathrm{resp}}=\ell_{\mathrm{pos}}+\lambda_{\mathrm{vel}}\ell_{\mathrm{vel}}+\lambda_{\mathrm{end}}\ell_{\mathrm{end}}+\lambda_{\mathrm{valid}}\ell_{\mathrm{valid}}.
\label{eq:main_response_loss}
\end{equation}
The position and velocity terms compare the agent state estimates with valid reference states, while the endpoint term supervises each trajectory's last valid position.
The validity term supervises the estimated probability that each agent state is valid.
Appendix~\ref{app:response_supervision} defines the losses, while Appendix~\ref{app:kinematic_grounding} relates position and velocity errors to differences between estimated and reference kinematic residuals.

\begin{figure*}[t]
  \centering
  \includegraphics[width=.88\textwidth]{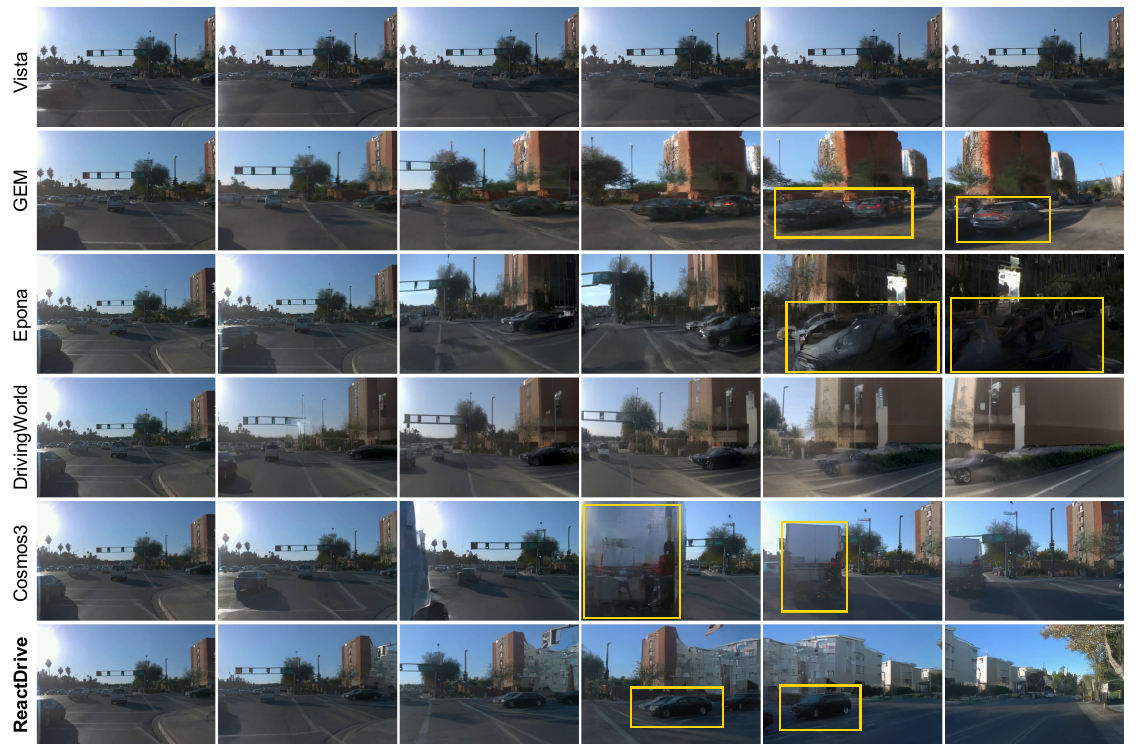}
  \vspace{-5pt}
  \caption{Visual comparison during a queried right turn.
  Rows show different models and columns show successive frames generated from the same observed history and ego query.
  Yellow boxes highlight differences in vehicle shape and the abrupt appearance of agents in the foreground.
  \reactdrive{} maintains recognizable surrounding vehicles as the generated viewpoint follows the turn.}
  \vspace{-10pt}
  \label{fig:qualitative}
\end{figure*}

\subsection{Dynamics-Conditioned Visual Generation}
\label{sec:visual_generation}

To incorporate agent responses into visual generation, the response encoder maps agent states to tokens that condition the video transformer.
For each future frame, it combines the estimated states of all $N$ agents into one response token:
\begin{equation}
\mathbf{c}_{t}=g_{\psi}([\widehat{m}_{t,1}\widehat{\mathbf{s}}_{t,1};\ldots;\widehat{m}_{t,N}\widehat{\mathbf{s}}_{t,N}])+\mathbf{e}_{\mathrm{resp}}.
\label{eq:main_response_tokens}
\end{equation}
Here, $g_{\psi}$ projects the weighted agent states into the video transformer's token space, and $\mathbf{e}_{\mathrm{resp}}$ identifies the response modality.
Validity weighting reduces the contribution of states with low estimated validity.
Agent states are concatenated in a fixed order across frames.
Applying the same encoder at each future time gives
\begin{equation}
\mathbf{C}=G_{\psi}(\widehat{\mathbf{S}})=(\mathbf{c}_{1},\ldots,\mathbf{c}_{T}).
\end{equation}
The resulting response tokens follow the temporal order of the agent states. As shown in Figure~\ref{fig:pipeline}c, response tokens are concatenated with noisy future video tokens and the shared visual and action tokens for joint attention. This connection allows the agent responses to guide video generation alongside the observed scene and ego actions.
We incorporate response tokens alongside visual history and ego actions to generate futures that reflect the modeled agent dynamics.
Let $\mathbf{z}$ denote clean future video latents and $\boldsymbol{\epsilon}\sim\mathcal{N}(\mathbf{0},\mathbf{I})$.
The flow interpolates between them:
\begin{equation}
\mathbf{z}_{\tau}=(1-\tau)\mathbf{z}+\tau\boldsymbol{\epsilon}, \qquad \tau\in[0,1].
\label{eq:reactdrive_flow}
\end{equation}
The variable $\tau$ indexes the generative flow and is distinct from the physical time in Equation~\eqref{eq:physical_kinematics}. The resulting flow field is conditioned on response tokens computed from the agent state estimates:
\begin{equation}
\widetilde{\mathbf{u}}(\mathbf{z}_{\tau},\tau)=\mathbf{u}_{\theta}(\mathbf{z}_{\tau},\tau;\mathbf{h},G_{\psi}(F_{\phi}(\mathbf{h}))).
\label{eq:reactdrive_composition}
\end{equation}
The ego query influences future observations through both direct action conditioning and the agent responses modeled by the dynamics prior.
Response tokens are not corrupted by flow noise during training or sampling.
Appendix~\ref{app:response_encoding} details the response encoder.

\subsection{Joint Learning and Simulation}
\label{sec:reactdrive_learning}

Agent dynamics and video generation play complementary roles on agent response modeling.
Reference agent trajectories supervise the agent state estimates, while the video objective trains the framework to use these states in generating future observations.
We jointly optimize these objectives:
\begin{equation}
\mathcal{L}=\mathbb{E}\!\left[\frac{1}{D}\left\|\widetilde{\mathbf{u}}(\mathbf{z}_{\tau},\tau)-(\boldsymbol{\epsilon}-\mathbf{z})\right\|_{2}^{2}+\lambda_{\mathrm{resp}}\ell_{\mathrm{resp}}\right].
\label{eq:reactdrive_objective}
\end{equation}
The first term is the rectified-flow objective over the $D$ scalar entries of the video latents.
The second supervises the state estimates, with $\lambda_{\mathrm{resp}}$ balancing the two objectives.

Gradients from the video objective propagate through the response encoder $G_{\psi}$ into the dynamics module $F_{\phi}$. Since validity probabilities enter the response tokens as continuous weights, these gradients reach the estimated positions, velocities, and validity probabilities. Dynamics supervision grounds these estimates in reference agent dynamics, while the video objective trains them to provide useful conditions for future observations. During joint training, the generator receives response tokens computed from the model's own state estimates, using the same source of conditions as during simulation. Appendix~\ref{app:optimization} specifies the optimization and training configuration.

During simulation,
\reactdrive{} uses the learned dynamics prior to compute agent state estimates from observed history and the queried ego trajectory, then encodes them into response tokens.
These conditions remain fixed while the learned flow transforms Gaussian noise into future video latents, which are decoded into visual observations.
For each alternative ego query, \reactdrive{} recomputes the agent state estimates and response tokens before generating a corresponding visual rollout from the same observed scene.
Our proposed framework requires no prescribed future agent trajectories, as demonstrated in Appendix~\ref{app:inference}.

\begin{table*}[t]
\centering
\fontsize{8}{9.2}\selectfont
\setlength{\tabcolsep}{1.8pt}
\renewcommand{\arraystretch}{1.08}

\resizebox{.9\linewidth}{!}{%
\begin{tabular}{ll*{6}{c}@{\hspace{7pt}}*{6}{c}}
\toprule
& & \multicolumn{6}{c}{\textbf{Waymo}} & \multicolumn{6}{c}{\textbf{nuPlan}} \\
\cmidrule(lr){3-8}\cmidrule(lr){9-14}
\textbf{Setting} & \textbf{Metric}
& Vista & GEM & Epona & DrivingWorld & Cosmos3 & \reactdrive{}
& Vista & GEM & Epona & DrivingWorld & Cosmos3 & \reactdrive{} \\
\midrule
\multicolumn{14}{c}{\textbf{Visual Quality and Self-reactivity}} \\
\midrule
\multirow{2}{*}{\textbf{Visual quality}}
& FID$\downarrow$ & 47.53 & 24.55 & 30.71 & 29.31 & \rankfirst{23.11} & \ranksecond{23.64} & 39.78 & 30.13 & 33.58 & 47.16 & \rankfirst{20.54} & \ranksecond{21.82} \\
& FVD$\downarrow$ & 42.78 & 12.99 & 15.38 & 13.67 & \ranksecond{9.91} & \rankfirst{7.30} & 48.66 & 14.30 & 9.64 & 23.54 & \ranksecond{8.66} & \rankfirst{6.93} \\
\midrule
\multirow{2}{*}{\textbf{Agent responses}}
& ADE$\downarrow$ & 2.09 & 1.58 & 1.88 & 1.60 & \ranksecond{1.14} & \rankfirst{1.03} & 1.47 & 1.39 & 1.47 & 2.07 & \ranksecond{0.98} & \rankfirst{0.73} \\
& FDE$\downarrow$ & 5.23 & 4.08 & 5.72 & 4.20 & \ranksecond{2.58} & \rankfirst{2.20} & 2.48 & 2.23 & 4.43 & 4.60 & \ranksecond{1.48} & \rankfirst{1.22} \\
\midrule
\multicolumn{14}{c}{\textbf{Ego Controllability and Collision Rate}} \\
\midrule
\multirow{3}{*}{\textbf{Overall}}
& ADE$\downarrow$ & 21.78 & 10.32 & 12.17 & 10.59 & \ranksecond{2.37} & \rankfirst{1.86} & 18.61 & 7.11 & 8.38 & 12.06 & \ranksecond{2.78} & \rankfirst{2.21} \\
& FDE$\downarrow$ & 40.07 & 21.14 & 20.49 & 21.15 & \ranksecond{4.64} & \rankfirst{3.49} & 34.76 & 14.13 & 16.29 & 23.92 & \ranksecond{5.29} & \rankfirst{3.91} \\
& CR (\%)$\downarrow$ & 2.14 & 0.87 & 3.06 & 1.89 & \ranksecond{0.47} & \rankfirst{0.26} & 0.63 & \ranksecond{0.41} & 3.10 & 1.20 & 0.53 & \rankfirst{0.09} \\
\midrule
\multirow{3}{*}{\textbf{Lane change}}
& ADE$\downarrow$ & 24.72 & 15.18 & 17.75 & 13.10 & \ranksecond{3.38} & \rankfirst{2.93} & 21.07 & 7.88 & 8.55 & 13.56 & \ranksecond{3.46} & \rankfirst{3.21} \\
& FDE$\downarrow$ & 46.96 & 29.53 & 27.22 & 26.58 & \ranksecond{6.16} & \rankfirst{5.24} & 40.78 & 14.08 & 15.10 & 28.47 & \ranksecond{6.18} & \rankfirst{5.22} \\
& CR (\%)$\downarrow$ & 1.59 & 1.07 & 4.18 & 2.49 & \ranksecond{0.62} & \rankfirst{0.48} & 0.39 & 0.72 & 4.51 & 1.22 & \ranksecond{0.23} & \rankfirst{0.20} \\
\midrule
\multirow{3}{*}{\textbf{Accel./decel.}}
& ADE$\downarrow$ & 24.73 & 7.03 & 7.06 & 11.90 & \ranksecond{2.11} & \rankfirst{1.33} & 21.28 & 6.64 & 7.06 & 14.08 & \ranksecond{2.73} & \rankfirst{1.98} \\
& FDE$\downarrow$ & 47.38 & 13.74 & 11.79 & 24.42 & \ranksecond{4.84} & \rankfirst{2.90} & 41.40 & 12.37 & 13.56 & 28.37 & \ranksecond{6.01} & \rankfirst{4.13} \\
& CR (\%)$\downarrow$ & 2.23 & \ranksecond{0.52} & 2.78 & 1.67 & \ranksecond{0.52} & \rankfirst{0.16} & 0.91 & \ranksecond{0.19} & 1.81 & 1.06 & 0.38 & \rankfirst{0.03} \\
\midrule
\multirow{3}{*}{\textbf{Stop}}
& ADE$\downarrow$ & 9.98 & 7.18 & 11.23 & 2.97 & \ranksecond{0.86} & \rankfirst{0.79} & 8.36 & 6.49 & 10.67 & 5.04 & \ranksecond{1.51} & \rankfirst{0.66} \\
& FDE$\downarrow$ & 11.67 & 19.17 & 24.41 & 3.73 & \ranksecond{1.20} & \rankfirst{1.19} & 9.45 & 17.76 & 24.13 & 5.94 & \ranksecond{2.06} & \rankfirst{0.84} \\
& CR (\%)$\downarrow$ & 3.06 & 1.18 & 1.39 & 1.11 & \ranksecond{0.07} & \rankfirst{0.00} & 0.56 & \ranksecond{0.25} & 2.88 & 1.44 & 1.44 & \rankfirst{0.00} \\
\bottomrule
\end{tabular}%
}

\caption{\textbf{Main results on Waymo and nuPlan.}
Under factual ego trajectories, \reactdrive{} achieves the lowest agent trajectory errors and FVD.
It follows counterfactual ego trajectories most accurately and has the lowest estimated collision rate. The best results are bolded and highlighted in blue, while the second-best results are underlined and highlighted in yellow. }
\vspace{-10pt}
\label{tab:main_results}
\label{tab:waymo_main}
\label{tab:nuplan_main}
\end{table*}

\section{Experiments}
\label{sec:experiments}

In this section, we comprehensively evaluate \reactdrive{} across multiple dimensions, including visual quality, ego controllability, \selfreactivity{}, and safety. Section~\ref{sec:experimental_setup} establishes the evaluation protocol, and Section~\ref{sec:main_results} examines generated futures under factual and counterfactual ego trajectories. We further conduct a human study in Section~\ref{sec:human_evaluation} to assess whether the \selfreactivity{}, ego controllability, and visual quality align with human expectations. To evaluate the practical benefits of simulated driving scenes, Section~\ref{sec:planning} investigates the impact of our generated videos on downstream planning performance. Finally, ablations studies are presented in Section~\ref{sec:ablation} to evaluate the effectiveness of the reactive dynamics prior and position information.

\subsection{Experimental Settings}
\label{sec:experimental_setup}

\myparagraph{Datasets and Baselines.}We train \reactdrive{} on the full training splits of the Waymo Open Dataset~\citep{sun2020scalability} and nuPlan~\citep{caesar2021nuplan}. Comparisons with Vista~\citep{gao2024vista}, GEM~\citep{hassan2025gem}, Epona~\citep{Zhang_2025_ICCV}, DrivingWorld~\citep{hu2026drivingworld}, and Cosmos3~\citep{agarwal2026cosmos} use the same observed history, specified ego actions, prediction horizon, and evaluation procedures. Each model observes 13 frames and generates 48 future frames at 10\,Hz. Downstream planning uses the official splits of NAVSIM~\citep{dauner2024navsim}. Appendix~\ref{app:method_details} provides the training details.

\myparagraph{Evaluation Metrics.}We report Fr\'echet Inception Distance (FID)~\citep{heusel2017gans} and Fr\'echet Video Distance (FVD)~\citep{unterthiner2018towards} for visual quality, average and final displacement errors (ADE/FDE) for trajectory accuracy, and Collision Rate (CR)~\citep{zeng2019end} estimated from vehicle detections in generated videos. Planning metrics follow NAVSIM~\citep{dauner2024navsim}: Predictive Driver Model Score (PDMS), no-at-fault collisions (NC), drivable-area compliance (DAC), compliance with the time-to-collision bound (TTC), ego progress (EP), dynamic comfort, and driving-direction compliance (DDC). Appendices~\ref{app:evaluation_metrics} and~\ref{app:navsim_evaluation} provide the metric definitions.

\subsection{Main Results}
\label{sec:main_results}

To evaluate \reactdrive's ability to reproduce observed agent dynamics and simulate plausible futures under alternative ego actions, we conduct experiments using factual and counterfactual ego trajectories, respectively. Factual ego trajectories provide reference videos for FID and FVD and reference agent trajectories for normalized image-space Agent ADE/FDE. Under edited ego trajectories, Ego ADE/FDE measure ego controllability, while CR measures the estimated collision rate. We apply five maneuvers to the same 100 test scenes per dataset, yielding 500 videos per model. The overall results weight left and right lane changes, acceleration, deceleration, and stopping equally.

\myparagraph{Self-reactivity.}\reactdrive{} achieves the lowest Agent ADE/FDE and FVD on both datasets in Table~\ref{tab:main_results}.
Relative to Cosmos3, Agent FDE decreases from 2.58 to 2.20 on Waymo and from 1.48 to 1.22 on nuPlan.
FVD decreases by 26.3\% and 20.0\%, respectively, while Cosmos3 retains the lowest FID.
Agent ADE/FDE compare matched agent tracks in the generated and recorded videos. The lower errors show that, under factual ego trajectories, agents in \reactdrive{} videos more closely follow their observed trajectories.
On both datasets, the lower agent errors accompany a closer match between generated and recorded clips in video feature space.
Section~\ref{sec:human_evaluation} examines the complementary question of whether the generated agent responses appear plausible when the ego action is edited.

\myparagraph{Ego Controllability and Collision Rate.}\reactdrive{} achieves the lowest Ego ADE/FDE and CR in every reported action group on both datasets. Compared with Cosmos3, overall Ego ADE decreases from 2.37 to 1.86 on Waymo and from 2.78 to 2.21 on nuPlan, while CR decreases from 0.47\% to 0.26\% and from 0.53\% to 0.09\%. Ego controllability is essential for assessing \selfreactivity{}, since surrounding agent responses need to be evaluated in relation to the action actually executed by the ego vehicle. Across the counterfactual maneuvers, \reactdrive{} follows the requested ego trajectory more closely while fewer generated frames are flagged for a potential collision.

\begin{table}[t]
\centering
\setlength{\tabcolsep}{2pt}
\renewcommand{\arraystretch}{1.05}
\resizebox{.95\linewidth}{!}{\begin{tabular}{@{}lcccc}
\toprule
\textbf{Method}
& \textbf{\shortstack{Agent\\reactivity}}
& \textbf{\shortstack{Ego\\controllability}}
& \textbf{\shortstack{Visual\\quality}}
& \textbf{\shortstack{Overall\\preference}} \\
\midrule
\reactdrive{}
& \rankfirst{598 (96.0\%)}
& \rankfirst{540 (86.7\%)}
& \rankfirst{572 (91.8\%)}
& \rankfirst{611 (98.1\%)} \\
Baselines
& \ranksecond{25 (4.0\%)}
& \ranksecond{83 (13.3\%)}
& \ranksecond{51 (8.2\%)}
& \ranksecond{12 (1.9\%)} \\
\bottomrule
\end{tabular}
}
\caption{\textbf{Human evaluation.}
Under counterfactual ego actions, \reactdrive{} is preferred for self-reactivity, ego controllability, and visual quality, with the strongest preference for its videos overall.}
\label{tab:human_evaluation}
\vspace{-10pt}
\end{table}

\myparagraph{Qualitative Comparison.}Figure~\ref{fig:qualitative} compares generated sequences under the same specified ego right turn. Vista exhibits limited viewpoint change, while GEM and Epona show degraded agent appearance and scene structure. Cosmos3 exhibits the abrupt appearance of a truck in the immediate foreground. \reactdrive{} preserves recognizable surrounding agents and scene structure throughout the turn.
Distorted agents and abrupt appearances obscure how surrounding agents respond as the ego action unfolds. By preserving recognizable agents throughout the turn, \reactdrive{} provides a clearer visual representation of their behavior under the specified maneuver. 
For additional visualizations, please refer to Appendix~\ref{app:visualization}.

\subsection{Human Evaluation}
\label{sec:human_evaluation}

Since agent responses to counterfactual ego actions are unobserved in recorded data, we conduct a human study to evaluate \selfreactivity{} by assessing the plausibility of generated agent responses under specified maneuvers, alongside ego controllability and visual quality. Specifically, we perform blinded pairwise comparisons using the same observed history and ego action. The study yields 623 comparisons per criterion, pooled across five baselines, with details provided in Appendix~\ref{app:human_evaluation}.

\myparagraph{Self-reactivity.}\reactdrive{} is preferred for agent reactivity in 96.0\% of the comparisons in Table~\ref{tab:human_evaluation}. Each comparison uses the same observed history and specified ego action, allowing participants to assess agent responses in the context of the same maneuver. The preference supports the plausibility of \reactdrive{}'s responses under ego actions. Together with the agent errors measured under factual ego trajectories, these judgments provide complementary evidence for \selfreactivity{} across evaluation settings.

\myparagraph{Ego Controllability and Visual Quality.}\reactdrive{} is preferred for ego controllability in 86.7\% of comparisons and visual quality in 91.8\%. These criteria assess the execution of the specified ego action and the rendering of the resulting scene. Both contribute to interpreting agent responses: ego controllability establishes the maneuver to which agents respond, while visual quality keeps those responses recognizable throughout the sequence. Across the same video pairs, participants favored \reactdrive{} more often for agent reactivity than for visual quality. In some comparisons, they therefore preferred the visual quality of a baseline but judged the surrounding agents' responses more appropriate in \reactdrive{}.

\myparagraph{Overall Preference.} \reactdrive{} receives 98.1\% overall preference when participants assess the generated sequence as a whole. This criterion allows viewers to consider the executed ego action, surrounding agent responses, and scene appearance together. The overall preference shows that favorable judgments of individual properties also extend to the complete generated sequence. Together with the agent reactivity results, this assessment supports the perceived quality of \reactdrive{} when agent responses, ego actions, and scene appearance are considered jointly.

\begin{table}[t]
\centering
\fontsize{8}{9.2}\selectfont
\resizebox{.9\linewidth}{!}{
\begin{tabular}{@{}lccc}
\toprule
\textbf{Metric} & \textbf{Real data} & \textbf{$+$Cosmos3} & \textbf{$+$\reactdrive{}} \\
\midrule
NC (\%)$\uparrow$ & \ranksecond{97.8} & \ranksecond{97.8} & \rankfirst{98.4} \\
DAC (\%)$\uparrow$ & \ranksecond{93.0} & \ranksecond{93.0} & \rankfirst{93.8} \\
TTC (\%)$\uparrow$ & \ranksecond{93.8} & 93.4 & \rankfirst{94.4} \\
EP (\%)$\uparrow$ & 77.8 & \ranksecond{77.9} & \rankfirst{78.7} \\
Comfort (\%)$\uparrow$ & \rankfirst{100.0} & \rankfirst{100.0} & \rankfirst{100.0} \\
DDC (\%)$\uparrow$ & \ranksecond{98.3} & \ranksecond{98.3} & \rankfirst{98.4} \\
\midrule
\textbf{PDMS$\uparrow$} & \ranksecond{83.8} & \ranksecond{83.8} & \rankfirst{85.1} \\
\bottomrule
\end{tabular}}
\caption{\textbf{Downstream task.}
On NAVSIM, adding \reactdrive{} rollouts to planner training yields the highest overall planning score, with gains in both driving safety and efficiency.}
\vspace{-10pt}
\label{tab:navsim_downstream}
\end{table}

\begin{table*}[t]
\centering
\setlength{\tabcolsep}{3pt}
\renewcommand{\arraystretch}{1.06}
\resizebox{.88\textwidth}{!}{%
\begin{tabular}{@{}l*{16}{c}}
\toprule
& \multicolumn{4}{c}{\textbf{Factual ego trajectories}} & \multicolumn{12}{c}{\textbf{Counterfactual ego trajectories}} \\
\cmidrule(lr){2-5}\cmidrule(lr){6-17}
& \multicolumn{2}{c}{Visual quality} & \multicolumn{2}{c}{Agent responses} & \multicolumn{3}{c}{Overall} & \multicolumn{3}{c}{Lane change} & \multicolumn{3}{c}{Accel./decel.} & \multicolumn{3}{c}{Stop} \\
\cmidrule(lr){2-3}\cmidrule(lr){4-5}\cmidrule(lr){6-8}\cmidrule(lr){9-11}\cmidrule(lr){12-14}\cmidrule(lr){15-17}
\textbf{Method} & FID$\downarrow$ & FVD$\downarrow$ & ADE$\downarrow$ & FDE$\downarrow$ & ADE$\downarrow$ & FDE$\downarrow$ & CR (\%)$\downarrow$ & ADE$\downarrow$ & FDE$\downarrow$ & CR (\%)$\downarrow$ & ADE$\downarrow$ & FDE$\downarrow$ & CR (\%)$\downarrow$ & ADE$\downarrow$ & FDE$\downarrow$ & CR (\%)$\downarrow$ \\
\midrule
w/o Dynamics Prior & \rankfirst{23.47} & 7.42 & 1.14 & \ranksecond{2.27} & \ranksecond{2.19} & 4.11 & 0.45 & \ranksecond{3.39} & 6.08 & 0.90 & \ranksecond{1.48} & 3.19 & \ranksecond{0.20} & \ranksecond{1.21} & 2.00 & \ranksecond{0.07} \\
w/o Position & \ranksecond{23.57} & \ranksecond{7.33} & \ranksecond{1.07} & 3.02 & 7.01 & \ranksecond{3.68} & \ranksecond{0.33} & 8.48 & \ranksecond{5.67} & \ranksecond{0.55} & 7.11 & \ranksecond{2.91} & 0.23 & 3.85 & \ranksecond{1.22} & \ranksecond{0.07} \\
\reactdrive{} & 23.64 & \rankfirst{7.30} & \rankfirst{1.03} & \rankfirst{2.20} & \rankfirst{1.86} & \rankfirst{3.49} & \rankfirst{0.26} & \rankfirst{2.93} & \rankfirst{5.24} & \rankfirst{0.48} & \rankfirst{1.33} & \rankfirst{2.90} & \rankfirst{0.16} & \rankfirst{0.79} & \rankfirst{1.19} & \rankfirst{0.00} \\
\bottomrule
\end{tabular}}
\caption{\textbf{Ablation study on Waymo.}
We compare \reactdrive{} with variants without the reactive dynamics prior or position information, evaluating agent responses, ego controllability, visual quality, and collision rate.}
\vspace{-10pt}
\label{tab:ablation_waymo}
\end{table*}

\subsection{Downstream Task}
\label{sec:planning}

To evaluate the practical impact of \reactdrive{}, we investigate whether its generated videos can improve downstream planning performance. Specifically, we train TransFuser~\citep{chitta2022transfuser} planners on NAVSIM~\citep{dauner2024navsim}, augmenting the training data with equal numbers of driving scenes simulated by \reactdrive{} and Cosmos3~\citep{agarwal2026cosmos}, respectively. Both simulators provide front-camera observations under the same data construction and planner training protocol, detailed in Appendix~\ref{app:navsim_evaluation}.

\myparagraph{Planning Performance.}Table~\ref{tab:navsim_downstream} reports a PDMS of 85.1 with \reactdrive{} augmentation, compared with 83.8 for both real-data training and Cosmos3 augmentation. The two generators contribute equal numbers of observations under the same planner training protocol. With the source scenes, ego trajectory labels, and planner training protocol held fixed, the higher PDMS shows that the front-camera observations generated by \reactdrive{} provide more effective training examples for TransFuser. The improvement thus extends from the generated videos to the performance of the planner trained on them.

\myparagraph{Collision Avoidance.}Relative to Cosmos3 augmentation, \reactdrive{} improves NC from 97.8 to 98.4 and TTC from 93.4 to 94.4. NC evaluates at-fault collisions during the simulated rollout, while TTC assesses compliance with the time-to-collision bound. These criteria capture complementary aspects of collision avoidance, including interactions with surrounding agents. The improvement across both scores shows that the benefit of \reactdrive{} augmentation extends to collision avoidance by the trained planner. This result connects the generated experience to downstream driving behavior and complements the evaluations of agent responses within generated videos.

\myparagraph{Driving Safety and Efficiency.}Beyond collision avoidance, the planner is evaluated on whether it stays within the drivable area and follows the driving direction. Relative to Cosmos3 augmentation, \reactdrive{} improves DAC from 93.0 to 93.8 and DDC from 98.3 to 98.4, while Comfort remains at 100.0. EP also increases from 77.9 to 78.7, measuring the distance traveled along the route relative to a reference planner. The improvement in EP indicates that better compliance with road constraints accompanies a greater distance traveled. Together with the higher NC and TTC scores, these results show that \reactdrive{} augmentation benefits several aspects of the planner's driving behavior.

\subsection{Ablation Study}
\label{sec:ablation}

The reactive dynamics prior supplies agent responses to visual generation, while position information preserves the spatial and temporal structure of the inputs used to model those responses. The ablations in Table~\ref{tab:ablation_waymo} examine their contributions to self-reactivity, ego controllability, and visual quality. The variant without the reactive dynamics prior removes the dynamics module and response encoder, leaving video generation conditioned only on visual history and ego actions. The second variant removes spatial and temporal position information from the visual and action tokens supplied to the dynamics module, as detailed in Appendix~\ref{app:response_predictor}.

\myparagraph{Reactive Dynamics Prior.} Removing the reactive dynamics prior increases Agent ADE/FDE under factual ego trajectories and Ego ADE/FDE under counterfactual ego trajectories. Overall CR also increases from 0.26\% to 0.45\%, with the largest deterioration for lane changes, where it rises from 0.48\% to 0.90\%. Both models generate video from the same visual history and ego action. When the generator also conditions on its own estimates of agent states, the tracks recovered from its videos more closely follow the surrounding agents' factual trajectories.

\myparagraph{Position Information.}Without position information, overall Ego ADE increases from 1.86 to 7.01, while Ego FDE changes from 3.49 to 3.68. The much larger increase in Ego ADE indicates greater deviations along the ego trajectory despite a relatively small change in the endpoint error. Agent FDE under factual ego trajectories also increases from 2.20 to 3.02. Position information identifies where visual features originate and when visual observations and ego actions occur. The degradation after removing this information from the dynamics module shows that organizing these inputs contributes to both the represented self-reactivity and ego controllability in generated videos.

\myparagraph{Limitations of Visual Metrics.} Across the model comparisons and ablations, FID ranges from 23.47 to 23.64 and FVD from 7.30 to 7.42. These closely grouped visual scores accompany substantial differences in agent errors, ego controllability, and collision rate. Ranking these variants by FID alone would favor the model without the dynamics prior, despite its higher agent and ego errors. The variant favored by FID therefore reproduces factual agent trajectories less accurately and follows counterfactual ego trajectories less closely. This reversal illustrates why evaluating an interactive world model requires separate measurements of visual quality, self-reactivity, and ego controllability.

\section{Conclusion}
\label{sec:conclusion}

Simulating an interactive world requires modeling the dynamics through which agents influence one another and the world evolves under their actions.
We define \selfreactivity{} to formalize this intrinsic requirement, making agent responses an explicit part of what the world model learns to simulate.
To realize this capability, we propose \reactdrive{}, a self-reactive driving world model that jointly learns agent dynamics and visual generation.
The reactive dynamics prior grounds agent responses in observed dynamics and uses the inferred responses to guide visual generation.
Joint dynamics and video supervision enable the framework to generate future scenes that reflect these responses under specified ego actions, without prescribed future agent trajectories.
Our experimental studies demonstrate more plausible agent reactions and downstream planning gains from augmentation with \reactdrive{} videos.
Our work highlights \selfreactivity{} as a fundamental capability for advancing world models from visually realistic generators toward faithful simulators of interactive environments.

\section*{LLM Usage Statement}

GPT assisted with language revision, \LaTeX{} preparation, figure presentation, and visualization code. It also supported consistency checks across the manuscript, citation checks, and analytical calculations of parameter counts and memory usage, using information and assumptions supplied by the authors. The research questions and methodological contributions originated with the authors. GPT was not used to perform model training or experimental evaluation, or to produce experimental measurements.

{
    \small
    \bibliographystyle{ieeenat_fullname}
    \bibliography{main}

@String(CVPR= {IEEE Conf. Comput. Vis. Pattern Recog.})

@String(ICCV= {Int. Conf. Comput. Vis.})

@String(AAAI = {AAAI})

@String(CVPR  = {CVPR})

@String(ICCV  = {ICCV})

@article{gao2024vista,
  title={Vista: A Generalizable Driving World Model with High Fidelity and Versatile Controllability},
  author={Gao, Shenyuan and Yang, Jiazhi and Chen, Li and Chitta, Kashyap and Qiu, Yihang and Geiger, Andreas and Zhang, Jun and Li, Hongyang},
  journal={arXiv preprint arXiv:2405.17398},
  year={2024}
}

@article{gao2023magicdrive,
  title={Magicdrive: Street view generation with diverse 3d geometry control},
  author={Gao, Ruiyuan and Chen, Kai and Xie, Enze and Hong, Lanqing and Li, Zhenguo and Yeung, Dit-Yan and Xu, Qiang},
  journal={arXiv preprint arXiv:2310.02601},
  year={2023}
}

@inproceedings{liu2024mvpbev,
  title={MVPbev: Multi-view Perspective Image Generation from BEV with Test-time Controllability and Generalizability},
  author={Liu, Buyu and Wang, Kai and Liu, Yansong and Bao, Jun and Han, Tingting and Yu, Jun},
  booktitle={Proceedings of the 32nd ACM International Conference on Multimedia},
  pages={8393--8401},
  year={2024}
}

@article{gao2024magicdrive3d,
  title={MagicDrive3D: Controllable 3D Generation for Any-View Rendering in Street Scenes},
  author={Gao, Ruiyuan and Chen, Kai and Li, Zhihao and Hong, Lanqing and Li, Zhenguo and Xu, Qiang},
  journal={arXiv preprint arXiv:2405.14475},
  year={2024}
}

@inproceedings{jiang2023motiondiffuser,
  title={Motiondiffuser: Controllable multi-agent motion prediction using diffusion},
  author={Jiang, Chiyu and Cornman, Andre and Park, Cheolho and Sapp, Benjamin and Zhou, Yin and Anguelov, Dragomir and others},
  booktitle={Proceedings of the IEEE/CVF Conference on Computer Vision and Pattern Recognition},
  year={2023}
}

@article{yang2023bevcontrol,
  title={Bevcontrol: Accurately controlling street-view elements with multi-perspective consistency via bev sketch layout},
  author={Yang, Kairui and Ma, Enhui and Peng, Jibin and Guo, Qing and Lin, Di and Yu, Kaicheng},
  journal={arXiv preprint arXiv:2308.01661},
  year={2023}
}

@article{swerdlow2024BEVGen,
  title={Street-view image generation from a bird's-eye view layout},
  author={Swerdlow, Alexander and Xu, Runsheng and Zhou, Bolei},
  journal={IEEE Robotics and Automation Letters},
  year={2024},
}

@inproceedings{wen2024panacea,
  title={Panacea: Panoramic and controllable video generation for autonomous driving},
  author={Wen, Yuqing and Zhao, Yucheng and Liu, Yingfei and Jia, Fan and Wang, Yanhui and Luo, Chong and Zhang, Chi and Wang, Tiancai and Sun, Xiaoyan and Zhang, Xiangyu},
  booktitle={Proceedings of the IEEE/CVF Conference on Computer Vision and Pattern Recognition},
  year={2024}
}

@article{yang2024drivearena,
  title={Drivearena: A closed-loop generative simulation platform for autonomous driving},
  author={Yang, Xuemeng and Wen, Licheng and Ma, Yukai and Mei, Jianbiao and Li, Xin and Wei, Tiantian and Lei, Wenjie and Fu, Daocheng and Cai, Pinlong and Dou, Min and others},
  journal={arXiv preprint arXiv:2408.00415},
  year={2024}
}

@article{heusel2017gans,
  title={Gans trained by a two time-scale update rule converge to a local nash equilibrium},
  author={Heusel, Martin and Ramsauer, Hubert and Unterthiner, Thomas and Nessler, Bernhard and Hochreiter, Sepp},
  journal={Advances in neural information processing systems},
  volume={30},
  year={2017}
}

@article{unterthiner2018towards,
  title={Towards accurate generative models of video: A new metric \& challenges},
  author={Unterthiner, Thomas and Van Steenkiste, Sjoerd and Kurach, Karol and Marinier, Raphael and Michalski, Marcin and Gelly, Sylvain},
  journal={arXiv preprint arXiv:1812.01717},
  year={2018}
}

@inproceedings{li2025uniscene,
  title={Uniscene: Unified occupancy-centric driving scene generation},
  author={Li, Bohan and Guo, Jiazhe and Liu, Hongsi and Zou, Yingshuang and Ding, Yikang and Chen, Xiwu and Zhu, Hu and Tan, Feiyang and Zhang, Chi and Wang, Tiancai and others},
  booktitle={Proceedings of the Computer Vision and Pattern Recognition Conference},
  pages={11971--11981},
  year={2025}
}

@inproceedings{sun2020scalability,
  title={Scalability in perception for autonomous driving: Waymo open dataset},
  author={Sun, Pei and Kretzschmar, Henrik and Dotiwalla, Xerxes and Chouard, Aurelien and Patnaik, Vijaysai and Tsui, Paul and Guo, James and Zhou, Yin and Chai, Yuning and Caine, Benjamin and others},
  booktitle={Proceedings of the IEEE/CVF conference on computer vision and pattern recognition},
  pages={2446--2454},
  year={2020}
}

@article{caesar2021nuplan,
  title={nuplan: A closed-loop ml-based planning benchmark for autonomous vehicles},
  author={Caesar, Holger and Kabzan, Juraj and Tan, Kok Seang and Fong, Whye Kit and Wolff, Eric and Lang, Alex and Fletcher, Luke and Beijbom, Oscar and Omari, Sammy},
  journal={arXiv preprint arXiv:2106.11810},
  year={2021}
}

@inproceedings{zhao2025drivedreamer,
  title={Drivedreamer-2: Llm-enhanced world models for diverse driving video generation},
  author={Zhao, Guosheng and Wang, Xiaofeng and Zhu, Zheng and Chen, Xinze and Huang, Guan and Bao, Xiaoyi and Wang, Xingang},
  booktitle={Proceedings of the AAAI Conference on Artificial Intelligence},
  volume={39},
  number={10},
  pages={10412--10420},
  year={2025}
}

@article{chen2024omnire,
  title={Omnire: Omni urban scene reconstruction},
  author={Chen, Ziyu and Yang, Jiawei and Huang, Jiahui and de Lutio, Riccardo and Esturo, Janick Martinez and Ivanovic, Boris and Litany, Or and Gojcic, Zan and Fidler, Sanja and Pavone, Marco and others},
  journal={arXiv preprint arXiv:2408.16760},
  year={2024}
}

@InProceedings{yan2025streetcrafter,
      author       = {Yan, Yunzhi and Xu, Zhen and Lin, Haotong and Jin, Haian and Guo, Haoyu and Wang, Yida and Zhan, Kun and Lang, Xianpeng and Bao, Hujun and Zhou, Xiaowei and Peng, Sida},
      title        = {StreetCrafter: Street View Synthesis with Controllable Video Diffusion Models},
      booktitle    = {Proceedings of the IEEE/CVF Conference on Computer Vision and Pattern Recognition (CVPR)},
      year         = {2025},
      pages        = {822--832}
}

@InProceedings{ni2025recondreamer,
      author       = {Ni, Chaojun and Zhao, Guosheng and Wang, Xiaofeng and Zhu, Zheng and Qin, Wenkang and Huang, Guan and Liu, Chen and Chen, Yuyin and Wang, Yida and Zhang, Xueyang and Zhan, Yifei and Zhan, Kun and Jia, Peng and Lang, Xianpeng and Wang, Xingang and Mei, Wenjun},
      title        = {ReconDreamer: Crafting World Models for Driving Scene Reconstruction via Online Restoration},
      booktitle    = {Proceedings of the IEEE/CVF Conference on Computer Vision and Pattern Recognition (CVPR)},
      year         = {2025},
      pages        = {1559--1569}
}

@InProceedings{hassan2025gem,
      author       = {Hassan, Mariam and Stapf, Sebastian and Rahimi, Ahmad and Rezende, Pedro M. B. and Haghighi, Yasaman and Bruggemann, David and Katircioglu, Isinsu and Zhang, Lin and Chen, Xiaoran and Saha, Suman and others},
      title        = {{GEM}: A Generalizable Ego-Vision Multimodal World Model for Fine-Grained Ego-Motion, Object Dynamics, and Scene Composition Control},
      booktitle    = {Proceedings of the IEEE/CVF Conference on Computer Vision and Pattern Recognition (CVPR)},
      year         = {2025}
}

@Article{hu2023gaia1,
  author    = {Anthony Hu and Lloyd Russell and Hudson Yeo and Zak Murez and 
               George Fedoseev and Alex Kendall and Jamie Shotton and Gianluca Corrado},
  title     = {{GAIA-1}: A Generative World Model for Autonomous Driving},
  journal   = {CoRR},
  volume    = {abs/2309.17080},
  year      = {2023}
}

@InProceedings{kim2021drivegan,
  author    = {Seung Wook Kim and Jonah Philion and Antonio Torralba and Sanja Fidler},
  title     = {DriveGAN: Towards a Controllable High-Quality Neural Simulation},
  booktitle = {Proceedings of the IEEE/CVF Conference on Computer Vision and Pattern Recognition (CVPR)},
  year      = {2021},
  month     = {June},
  pages     = {5820--5829}
}

@inproceedings{lu2024wovogen,
  title={Wovogen: World volume-aware diffusion for controllable multi-camera driving scene generation},
  author={Lu, Jiachen and Huang, Ze and Yang, Zeyu and Zhang, Jiahui and Zhang, Li},
  booktitle={European conference on computer vision},
  pages={329--345},
  year={2024},
  organization={Springer}
}

@article{russell2025gaia,
  title={Gaia-2: A controllable multi-view generative world model for autonomous driving},
  author={Russell, Lloyd and Hu, Anthony and Bertoni, Lorenzo and Fedoseev, George and Shotton, Jamie and Arani, Elahe and Corrado, Gianluca},
  journal={arXiv preprint arXiv:2503.20523},
  year={2025}
}

@inproceedings{yang2025resim,
  author    = {Jiazhi Yang and Kashyap Chitta and Shenyuan Gao and Long Chen and Yuqian Shao and Xiaosong Jia and Hongyang Li and Andreas Geiger and Xiangyu Yue and Li Chen},
  title     = {{ReSim}: Reliable World Simulation for Autonomous Driving},
  booktitle = {Advances in Neural Information Processing Systems (NeurIPS)},
  year      = {2025}
}

@InProceedings{Zhang_2025_ICCV,
    author    = {Zhang, Kaiwen and Tang, Zhenyu and Hu, Xiaotao and Pan, Xingang and Guo, Xiaoyang and Liu, Yuan and Huang, Jingwei and Yuan, Li and Zhang, Qian and Long, Xiao-Xiao and Cao, Xun and Yin, Wei},
    title     = {Epona: Autoregressive Diffusion World Model for Autonomous Driving},
    booktitle = {Proceedings of the IEEE/CVF International Conference on Computer Vision (ICCV)},
    month     = {October},
    year      = {2025},
    pages     = {27220-27230}
}

@inproceedings{tan2025scenediffuser,
  author    = {Tan, Shuhan and Lambert, John and Jeon, Hong and Kulshrestha, Sakshum and Bai, Yijing and Luo, Jing and Anguelov, Dragomir and Tan, Mingxing and Jiang, Chiyu Max},
  title     = {{SceneDiffuser++}: City-Scale Traffic Simulation via a Generative World Model},
  booktitle = {Proceedings of the IEEE/CVF Conference on Computer Vision and Pattern Recognition (CVPR)},
  year      = {2025},
  pages     = {1570--1580}
}

@article{chang2025langtraj,
  title={LANGTRAJ: Diffusion Model and Dataset for Language-Conditioned Trajectory Simulation},
  author={Chang, Wei-Jer and Zhan, Wei and Tomizuka, Masayoshi and Chandraker, Manmohan and Pittaluga, Francesco},
  journal={arXiv preprint arXiv:2504.11521},
  year={2025}
}

@inproceedings{wei2024editable,
  title={Editable scene simulation for autonomous driving via collaborative llm-agents},
  author={Wei, Yuxi and Wang, Zi and Lu, Yifan and Xu, Chenxin and Liu, Changxing and Zhao, Hao and Chen, Siheng and Wang, Yanfeng},
  booktitle={Proceedings of the IEEE/CVF Conference on Computer Vision and Pattern Recognition},
  pages={15077--15087},
  year={2024}
}

@inproceedings{ost2021nsg,
  title={Neural Scene Graphs for Dynamic Scenes},
  author={Ost, Julian and Mannan, Fahim and Thuerey, Nils and Knodt, Julian and Heide, Felix},
  booktitle={Proceedings of the IEEE/CVF Conference on Computer Vision and Pattern Recognition},
  pages={2856--2865},
  year={2021}
}

@inproceedings{yang2023unisim,
  title={UniSim: A Neural Closed-Loop Sensor Simulator},
  author={Yang, Ze and Chen, Yun and Wang, Jingkang and Manivasagam, Sivabalan and Ma, Wei-Chiu and Yang, Anqi Joyce and Urtasun, Raquel},
  booktitle={Proceedings of the IEEE/CVF Conference on Computer Vision and Pattern Recognition},
  pages={1389--1399},
  year={2023}
}

@article{yang2023emernerf,
  title={EmerNeRF: Emergent Spatial-Temporal Scene Decomposition via Self-Supervision},
  author={Yang, Jiawei and Ivanovic, Boris and Litany, Or and Weng, Xinshuo and Kim, Seung Wook and Li, Boyi and Che, Tong and Xu, Danfei and Fidler, Sanja and Pavone, Marco and Wang, Yue},
  journal={arXiv preprint arXiv:2311.02077},
  year={2023}
}

@inproceedings{tonderski2024neurad,
  title={NeuRAD: Neural Rendering for Autonomous Driving},
  author={Tonderski, Adam and Lindstr{\"o}m, Carl and Hess, Georg and Ljungbergh, William and Svensson, Lennart and Petersson, Christoffer},
  booktitle={Proceedings of the IEEE/CVF Conference on Computer Vision and Pattern Recognition},
  pages={14895--14904},
  year={2024}
}

@article{park2025simsplat,
  title={SIMSplat: Predictive Driving Scene Editing with Language-aligned 4D Gaussian Splatting},
  author={Park, Sung-Yeon and Lee, Adam and Lu, Juanwu and Cui, Can and Jiang, Luyang and Gupta, Rohit and Han, Kyungtae and Moradipari, Ahmadreza and Wang, Ziran},
  journal={arXiv preprint arXiv:2510.02469},
  year={2025}
}

@inproceedings{lu2025infinicube,
  title={Infinicube: Unbounded and controllable dynamic 3d driving scene generation with world-guided video models},
  author={Lu, Yifan and Ren, Xuanchi and Yang, Jiawei and Shen, Tianchang and Wu, Zhangjie and Gao, Jun and Wang, Yue and Chen, Siheng and Chen, Mike and Fidler, Sanja and others},
  booktitle={Proceedings of the IEEE/CVF International Conference on Computer Vision},
  pages={27272--27283},
  year={2025}
}

@article{wang2024freevs,
  title={Freevs: Generative view synthesis on free driving trajectory},
  author={Wang, Qitai and Fan, Lue and Wang, Yuqi and Chen, Yuntao and Zhang, Zhaoxiang},
  journal={arXiv preprint arXiv:2410.18079},
  year={2024}
}

@inproceedings{tolstaya2021identifying,
  title={Identifying driver interactions via conditional behavior prediction},
  author={Tolstaya, Ekaterina and Mahjourian, Reza and Downey, Carlton and Varadarajan, Balakrishnan and Sapp, Benjamin and Anguelov, Dragomir},
  booktitle={2021 IEEE International Conference on Robotics and Automation (ICRA)},
  pages={3473--3479},
  year={2021},
  organization={IEEE}
}

@article{ngiam2021scene,
  title={Scene transformer: A unified architecture for predicting multiple agent trajectories},
  author={Ngiam, Jiquan and Caine, Benjamin and Vasudevan, Vijay and Zhang, Zhengdong and Chiang, Hao-Tien Lewis and Ling, Jeffrey and Roelofs, Rebecca and Bewley, Alex and Liu, Chenxi and Venugopal, Ashish and others},
  journal={arXiv preprint arXiv:2106.08417},
  year={2021}
}

@inproceedings{sun2022m2i,
  title={M2i: From factored marginal trajectory prediction to interactive prediction},
  author={Sun, Qiao and Huang, Xin and Gu, Junru and Williams, Brian C and Zhao, Hang},
  booktitle={2022 IEEE/CVF Conference on Computer Vision and Pattern Recognition (CVPR)},
  pages={6533--6542},
  year={2022},
  organization={IEEE}
}

@article{zhang2023trafficbots,
  title={Trafficbots: Towards world models for autonomous driving simulation and motion prediction},
  author={Zhang, Zhejun and Liniger, Alexander and Dai, Dengxin and Yu, Fisher and Van Gool, Luc},
  journal={arXiv preprint arXiv:2303.04116},
  year={2023}
}

@inproceedings{huang2023gameformer,
  title={Gameformer: Game-theoretic modeling and learning of transformer-based interactive prediction and planning for autonomous driving},
  author={Huang, Zhiyu and Liu, Haochen and Lv, Chen},
  booktitle={2023 IEEE/CVF International Conference on Computer Vision (ICCV)},
  pages={3880--3890},
  year={2023},
  organization={IEEE}
}

@article{zhou2024behaviorgpt,
  title={Behaviorgpt: Smart agent simulation for autonomous driving with next-patch prediction},
  author={Zhou, Zikang and Hu, Haibo and Chen, Xinhong and Wang, Jianping and Guan, Nan and Wu, Kui and Li, Yung-Hui and Huang, Yu-Kai and Xue, Chun J},
  journal={Advances in Neural Information Processing Systems},
  volume={37},
  pages={79597--79617},
  year={2024}
}

@article{wu2024smart,
  title={Smart: Scalable multi-agent real-time motion generation via next-token prediction},
  author={Wu, Wei and Feng, Xiaoxin and Gao, Ziyan and Kan, Yuheng},
  journal={Advances in Neural Information Processing Systems},
  volume={37},
  pages={114048--114071},
  year={2024}
}

@article{yan2026causaldrive,
  title={CausalDrive: Real-time Causal World Models for Autonomous Driving},
  author={Yan, Tianyi and Zheng, Huan and Chen, Dubing and Qu, Meizhi and Shen, Yingying and Zhou, Lijun and Tu, Mingfei and Wang, Bing and Chen, Guang and Ye, Hangjun and others},
  journal={arXiv preprint arXiv:2606.15341},
  year={2026}
}

@inproceedings{zhou2026drivinggen,
  title={Drivinggen: A comprehensive benchmark for generative video world models in autonomous driving},
  author={Zhou, Yang and Shao, Hao and Wang, Letian and Zong, Zhuofan and Li, Hongsheng and Waslander, Steven},
  booktitle={International Conference on Learning Representations},
  volume={2026},
  pages={103502--103524},
  year={2026}
}

@article{zhou2026toward,
  title={Toward Physically Consistent Driving Video World Models under Challenging Trajectories},
  author={Zhou, Jiawei and Zhu, Zhenxin and Du, Lingyi and Lyu, Linye and Zhou, Lijun and Wu, Zhanqian and Luo, Hongcheng and Tian, Zhuotao and Wang, Bing and Chen, Guang and others},
  journal={arXiv preprint arXiv:2603.24506},
  year={2026}
}

@article{wang2026horizonforge,
  title={Horizonforge: Driving scene editing with any trajectories and any vehicles},
  author={Wang, Yifan and Pittaluga, Francesco and Tasneem, Zaid and You, Chenyu and Chandraker, Manmohan and Jiang, Ziyu},
  journal={arXiv preprint arXiv:2602.21333},
  year={2026}
}

@inproceedings{hu2026drivingworld,
  title={Drivingworld: Constructing world model for autonomous driving via video gpt},
  author={Hu, Xiaotao and Jia, Mingkai and Guo, Xiaoyang and Zhang, Qian and Long, Xiao-xiao and Yin, Wei},
  booktitle={International Conference on Pattern Recognition},
  pages={276--291},
  year={2026},
  organization={Springer}
}

@article{chitta2022transfuser,
  title={Transfuser: Imitation with transformer-based sensor fusion for autonomous driving},
  author={Chitta, Kashyap and Prakash, Aditya and Jaeger, Bernhard and Yu, Zehao and Renz, Katrin and Geiger, Andreas},
  journal={IEEE transactions on pattern analysis and machine intelligence},
  volume={45},
  number={11},
  pages={12878--12895},
  year={2022},
  publisher={IEEE}
}

@article{agarwal2026cosmos,
  title={Cosmos 3: Omnimodal world models for physical ai},
  author={Agarwal, Niket and Ali, Arslan and Allen, Jon and Antolini, Martin and Aubame, Adeline and Azzolini, Alisson and Bai, Junjie and Bala, Maciej and Balaji, Yogesh and Bapst, Josh and others},
  journal={arXiv preprint arXiv:2606.02800},
  year={2026}
}

@article{arai2024act,
  title={Act-bench: Towards action controllable world models for autonomous driving},
  author={Arai, Hidehisa and Ishihara, Keishi and Takahashi, Tsubasa and Yamaguchi, Yu},
  journal={arXiv preprint arXiv:2412.05337},
  year={2024}
}

@inproceedings{zhu2026other,
  title={Other vehicle trajectories are also needed: A driving world model unifies ego-other vehicle trajectories in video latent space},
  author={Zhu, Jian and Jia, Zhengyu and Gao, Tian and Deng, Jiaxin and Li, Shidi and Zhang, Lang and Liu, Fu and Jia, Peng and Lang, Xianpeng},
  booktitle={Proceedings of the AAAI Conference on Artificial Intelligence},
  volume={40},
  number={16},
  pages={13934--13942},
  year={2026}
}

@article{ha2018world,
  title={World models},
  author={Ha, David and Schmidhuber, J{\"u}rgen},
  journal={arXiv preprint arXiv:1803.10122},
  volume={2},
  number={3},
  pages={440},
  year={2018}
}

@article{wang2026behaviorworldgen,
  title={BehaviorWorldGen: Closing the Loop between Action Models and World Simulators via Controllable Behavior-Aware Structured World Generation},
  author={Wang, Jiaqi and Zhang, Zhuo and Guan, Haining and Zhou, Tingguang and Cui, Haowen and Zhu, Zhongyang and Zheng, Yulong and Wang, ChuanYe and Chen, Xuefeng and Yang, Zhen and others},
  journal={arXiv preprint arXiv:2608.22187},
  year={2026}
}

@inproceedings{tian2026simscale,
  title={Simscale: Learning to drive via real-world simulation at scale},
  author={Tian, Haochen and Li, Tianyu and Liu, Haochen and Yang, Jiazhi and Qiu, Yihang and Li, Guang and Wang, Junli and Gao, Yinfeng and Zhang, Zhang and Wang, Liang and others},
  booktitle={Proceedings of the IEEE/CVF Conference on Computer Vision and Pattern Recognition},
  pages={36365--36374},
  year={2026}
}

@inproceedings{hafner2019learning,
  title={Learning latent dynamics for planning from pixels},
  author={Hafner, Danijar and Lillicrap, Timothy and Fischer, Ian and Villegas, Ruben and Ha, David and Lee, Honglak and Davidson, James},
  booktitle={International conference on machine learning},
  pages={2555--2565},
  year={2019},
  organization={PMLR}
}

@article{battaglia2016interaction,
  title={Interaction networks for learning about objects, relations and physics},
  author={Battaglia, Peter and Pascanu, Razvan and Lai, Matthew and Jimenez Rezende, Danilo and others},
  journal={Advances in neural information processing systems},
  volume={29},
  year={2016}
}

@inproceedings{li2024generative,
  title={Generative image dynamics},
  author={Li, Zhengqi and Tucker, Richard and Snavely, Noah and Holynski, Aleksander},
  booktitle={2024 IEEE/CVF Conference on Computer Vision and Pattern Recognition (CVPR)},
  pages={24142--24153},
  year={2024},
  organization={IEEE}
}

@InProceedings{pmlr-v267-zhou25t,
  title = 	 {{DINO}-{WM}: World Models on Pre-trained Visual Features enable Zero-shot Planning},
  author =       {Zhou, Gaoyue and Pan, Hengkai and Lecun, Yann and Pinto, Lerrel},
  booktitle = 	 {Proceedings of the 42nd International Conference on Machine Learning},
  pages = 	 {79115--79135},
  year = 	 {2025},
  editor = 	 {Singh, Aarti and Fazel, Maryam and Hsu, Daniel and Lacoste-Julien, Simon and Berkenkamp, Felix and Maharaj, Tegan and Wagstaff, Kiri and Zhu, Jerry},
  volume = 	 {267},
  series = 	 {Proceedings of Machine Learning Research},
  month = 	 {13--19 Jul},
  publisher =    {PMLR},
  url = 	 {https://proceedings.mlr.press/v267/zhou25t.html},
}

@InProceedings{Motamed_2026_WACV,
    author={Motamed, Saman and Culp, Laura and Swersky, Kevin and Jaini, Priyank and Geirhos, Robert},
    title={Do Generative Video Models Understand Physical Principles?},
    booktitle={Proceedings of the IEEE/CVF Winter Conference on Applications of Computer Vision (WACV)},
    year={2026}
}

@INPROCEEDINGS{Sadigh-RSS-16, 
    AUTHOR    = {Dorsa Sadigh AND Shankar Sastry AND Sanjit A. Seshia AND Anca D. Dragan}, 
    TITLE     = {Planning for Autonomous Cars that Leverage Effects on Human Actions}, 
    BOOKTITLE = {Proceedings of Robotics: Science and Systems}, 
    YEAR      = {2016}, 
    ADDRESS   = {AnnArbor, Michigan}, 
    MONTH     = {June}, 
    DOI       = {10.15607/RSS.2016.XII.029} 
}

@article{dauner2024navsim,
  title={Navsim: Data-driven non-reactive autonomous vehicle simulation and benchmarking},
  author={Dauner, Daniel and Hallgarten, Marcel and Li, Tianyu and Weng, Xinshuo and Huang, Zhiyu and Yang, Zetong and Li, Hongyang and Gilitschenski, Igor and Ivanovic, Boris and Pavone, Marco and others},
  journal={Advances in Neural Information Processing Systems},
  volume={37},
  pages={28706--28719},
  year={2024}
}

@InProceedings{Gao_2025_ICCV,
    author    = {Gao, Ruiyuan and Chen, Kai and Xiao, Bo and Hong, Lanqing and Li, Zhenguo and Xu, Qiang},
    title     = {MagicDrive-V2: High-Resolution Long Video Generation for Autonomous Driving with Adaptive Control},
    booktitle = {Proceedings of the IEEE/CVF International Conference on Computer Vision (ICCV)},
    month     = {October},
    year      = {2025},
    pages     = {28135-28144}
}

@article{hu2021lora,
  title={Lora: Low-rank adaptation of large language models},
  author={Hu, Edward J and Shen, Yelong and Wallis, Phillip and Allen-Zhu, Zeyuan and Li, Yuanzhi and Wang, Shean and Wang, Lu and Chen, Weizhu},
  journal={arXiv preprint arXiv:2106.09685},
  year={2021}
}

@inproceedings{zhou2019continuity,
  title={On the continuity of rotation representations in neural networks},
  author={Zhou, Yi and Barnes, Connelly and Lu, Jingwan and Yang, Jimei and Li, Hao},
  booktitle={2019 IEEE/CVF Conference on Computer Vision and Pattern Recognition (CVPR)},
  pages={5738--5746},
  year={2019},
  organization={IEEE}
}

@article{vaswani2017attention,
  title={Attention is all you need},
  author={Vaswani, Ashish and Shazeer, Noam and Parmar, Niki and Uszkoreit, Jakob and Jones, Llion and Gomez, Aidan N and Kaiser, {\L}ukasz and Polosukhin, Illia},
  journal={Advances in neural information processing systems},
  volume={30},
  year={2017}
}

@article{jocher2026ultralytics,
  title={Ultralytics YOLO26: unified real-time end-to-end vision models},
  author={Jocher, Glenn and Qiu, Jing and Liu, Mengyu and Lyu, Shuai and Akyon, Fatih Cagatay and Kalfaoglu, Muhammet Esat},
  journal={arXiv preprint arXiv:2606.03748},
  year={2026}
}

@article{kuhn1955hungarian,
  title={The Hungarian method for the assignment problem},
  author={Kuhn, Harold W},
  journal={Naval research logistics quarterly},
  volume={2},
  number={1-2},
  pages={83--97},
  year={1955},
  publisher={Wiley Online Library}
}

@inproceedings{wang2026vggtomega,
  title     = {{VGGT-$\Omega$}},
  author    = {Jianyuan Wang and Minghao Chen and Shangzhan Zhang and Nikita Karaev and Johannes Sch{\"o}nberger and Patrick Labatut and Piotr Bojanowski and David Novotny and Andrea Vedaldi and Christian Rupprecht},
  booktitle = {Proceedings of the IEEE/CVF Conference on Computer Vision and Pattern Recognition (CVPR)},
  year      = {2026}
}

@article{ma2026rewam,
  title={{ReWAM}: Reciprocal World Action Models for Interactive Autonomous Driving},
  author={Ma, Benshan and Liu, Pei and Zhong, Ruiguo and Zhang, Lang and Feng, Mingyue and Wang, Yaonong and Ma, Jun},
  journal={arXiv preprint arXiv:2609.39245},
  year={2026}
}

@book{kant2004metaphysical,
  title={Metaphysical foundations of natural science},
  author={Kant, Immanuel},
  year={2004},
  publisher={Cambridge University Press}
}

@inproceedings{zeng2019end,
  title={End-to-end interpretable neural motion planner},
  author={Zeng, Wenyuan and Luo, Wenjie and Suo, Simon and Sadat, Abbas and Yang, Bin and Casas, Sergio and Urtasun, Raquel},
  booktitle={2019 IEEE/CVF Conference on Computer Vision and Pattern Recognition (CVPR)},
  pages={8652--8661},
  year={2019},
  organization={IEEE}
}
}

% As in the supplied author kit, compile main + supplementary together so all
% cross-references resolve. Split the PDF before submission. The supplementary
% content is maintained in ONE file and starts on its own page.
\clearpage
\setcounter{page}{1}
\maketitlesupplementary

\appendix
% \section{Appendix}

\section{Method Details}
\label{app:method_details}

This section specifies the inputs, architecture, supervision, and optimization of \reactdrive{}.
We retain the notation of Section~\ref{sec:methods}: $\mathbf{X}$ denotes observed frames, $\mathbf{Y}$ future observations, $\mathbf{A}$ the ego action sequence, and $\widehat{\mathbf{S}}$ the physical agent state estimates.
The dynamics module $F_{\phi}$ computes physical agent states, the response encoder $G_{\psi}$ maps them to response tokens $\mathbf{C}$, and the transformer $\mathbf{u}_{\theta}$ models the generative flow conditioned on these tokens.

\subsection{Backbone and Input Representation}
\label{app:backbone_action}

\myparagraph{Video backbone.}\reactdrive{} uses the Cosmos3 Nano video generation pathway conditioned on ego actions~\citep{agarwal2026cosmos}.
Each example contains $H=13$ observed frames and $T=48$ future frames at 10\,Hz.
A frozen tokenizer encodes the complete training clip:
\begin{equation}
E_{\mathrm{tok}}([\mathbf{X};\mathbf{Y}])=[\mathbf{z}_{\mathrm{obs}};\mathbf{z}].
\label{eq:app_video_tokenization}
\end{equation}
Only the future latents $\mathbf{z}$ are corrupted during flow training.
The observed latents $\mathbf{z}_{\mathrm{obs}}$ depend only on the observed frames $\mathbf{X}$ and provide the clean visual input to the dynamics module.
The horizon $T$ counts frames; the number of latent frames is determined by the tokenizer's temporal compression.

\myparagraph{Ego actions.}Each action encodes the relative camera translation and rotation between consecutive frames:
\begin{equation}
\mathbf{a}_{t}=[\Delta\mathbf{p}_{t};\rho_{6\mathrm{D}}(\Delta\mathbf{R}_{t})]\in\mathbb{R}^{9}.
\label{eq:av_action}
\end{equation}
The translation has three components and is expressed in the local ego coordinate system for the corresponding action step.
The function $\rho_{6\mathrm{D}}$ maps the relative rotation to its continuous six-dimensional representation~\citep{zhou2019continuity}.
The observed ego trajectory and queried future trajectory form the complete action sequence $\mathbf{A}$ passed to the action encoder.

\myparagraph{Agent coordinates.}All agent positions and velocities are expressed on the ground plane in a coordinate system fixed at the last observed ego pose.
This reference is shared across agents and future times, including when the queried ego trajectory changes.
The agent velocity $\mathbf{v}_{t,n}$ describes the rate of change of agent position in this reference frame.
The flow velocity $\mathbf{u}_{\theta}$ describes a transformation of video latents along $\tau$.

\subsection{Reactive Dynamics Prior}
\label{app:response_predictor}

\myparagraph{Visual and action memory.}Let $E_{\mathrm{vid}}$ denote the frozen video patch projection and $E_{\mathrm{act}}$ the frozen ego-action encoder.
The Video Encoder in Figure~\ref{fig:pipeline}(a) comprises the frozen video tokenizer and patch projection.
The resulting visual and action token streams are shared by the dynamics module and the video transformer, with a common feature dimension $d_{\mathrm{pred}}$.
We add a spatiotemporal encoding $\mathbf{P}_{\mathrm{st}}$ to the observed visual tokens and a temporal encoding $\mathbf{P}_{\mathrm{t}}$ to the action tokens:
\begin{equation}
\mathbf{M}=[E_{\mathrm{vid}}(\mathbf{z}_{\mathrm{obs}})+\mathbf{P}_{\mathrm{st}};E_{\mathrm{act}}(\mathbf{A})+\mathbf{P}_{\mathrm{t}}].
\label{eq:app_predictor_memory}
\end{equation}
Concatenation is along the token dimension.
The visual encoding identifies temporal, height, and width coordinates in the observed token grid.
The action encoding identifies the times of the observed and queried ego actions.

\myparagraph{Agent state queries.}The dynamics module uses $N=16$ learned agent state queries $\mathbf{Q}\in\mathbb{R}^{N\times d_{\mathrm{pred}}}$, an eight-head cross-attention layer~\citep{vaswani2017attention}, and a state decoder implemented as a two-layer MLP $h_{\phi}$:
\begin{equation}
\begin{aligned}
\mathbf{R}&=\operatorname{CrossAttn}_{\phi}(\mathbf{Q},\mathbf{M},\mathbf{M}), \\
(\widehat{\mathbf{P}},\widehat{\mathbf{V}},\widehat{\mathbf{L}})&=\operatorname{reshape}(h_{\phi}(\mathbf{R})).
\end{aligned}
\label{eq:response_predictor}
\end{equation}
For each query, the state decoder outputs $5T$ values: two position components, two velocity components, and one validity logit for every future frame.
Thus, $\widehat{\mathbf{P}},\widehat{\mathbf{V}}\in\mathbb{R}^{T\times N\times2}$ and $\widehat{\mathbf{L}}\in\mathbb{R}^{T\times N}$.
We obtain $\widehat{m}_{t,n}=\sigma(\widehat{\ell}_{t,n})$ and assemble $\widehat{\mathbf{s}}_{t,n}=[\widehat{\mathbf{p}}_{t,n};\widehat{\mathbf{v}}_{t,n}]$.
These operations define $F_{\phi}$.
For a fixed visual history and ego action, the dynamics module produces a deterministic sequence of physical agent states.
Each validity probability estimates whether the corresponding agent state is valid.

\myparagraph{Reference track assignment.}At the last observed frame, up to 16 reference tracks are assigned to the agent state queries, selecting the nearest valid agents in the front camera half plane.
Each assigned query retains the same reference agent throughout the rollout.
Unused slots and temporarily invalid states are padded.
Position and velocity supervision use the valid frames of each reference track. Endpoint supervision uses its last valid frame.
Reference future trajectories provide dynamics supervision, while the dynamics module receives only observed video latents and ego actions.

\subsection{Response Encoding}
\label{app:response_encoding}

Object states provide a basis for modeling physical interactions~\citep{battaglia2016interaction}, and dynamics learned from video can guide visual synthesis~\citep{li2024generative}.
The response encoder uses a fixed agent order across frames and produces one response token for each future frame.
For each future time, it weights the physical state of each agent by its estimated validity probability:
\begin{equation}
\widetilde{\mathbf{s}}_{t,n}=\widehat{m}_{t,n}\widehat{\mathbf{s}}_{t,n}, \qquad \mathbf{c}_{t}=g_{\psi}([\widetilde{\mathbf{s}}_{t,1};\ldots;\widetilde{\mathbf{s}}_{t,N}])+\mathbf{e}_{\mathrm{resp}}.
\label{eq:trajectory_conditioning}
\end{equation}
The projection $g_{\psi}:\mathbb{R}^{4N}\rightarrow\mathbb{R}^{d}$ and modality embedding $\mathbf{e}_{\mathrm{resp}}\in\mathbb{R}^{d}$ share the parameter group $\psi$.
For $N=16$, the projection receives 64 weighted state components per frame.
The same projection is applied at each future time, producing $T=48$ response tokens.
The sequence map $G_{\psi}$ collects these operations so that $G_{\psi}(\widehat{\mathbf{S}})=\mathbf{C}=(\mathbf{c}_{1},\ldots,\mathbf{c}_{T})$.

The response tokens are appended to the Cosmos3 input sequence and participate in joint attention with noisy future video tokens and the shared visual and action tokens.
Their indices follow future frame times; the video tokens use the temporal resolution of the tokenizer.
Response tokens remain clean during training and sampling and are excluded from the flow prediction target.
During simulation, the entire response token sequence is computed once for each ego query and reused at every flow step.

\subsection{Dynamics Supervision}
\label{app:response_supervision}

Reference tracks provide positions $\mathbf{p}_{t,n}$, velocities $\mathbf{v}_{t,n}$, and validity labels $m_{t,n}$.
We use $D_{p}=\operatorname{diag}(100,50)$ and $D_{v}=\operatorname{diag}(30,30)$ to normalize position and velocity errors.
These scales correspond to the channels $(x,y,v^{x},v^{y})$.
Writing $M=\sum_{t,n}m_{t,n}$ for the number of valid future states, the position and velocity losses for $M>0$ are
\begin{equation}
\ell_{\mathrm{pos}}=\frac{1}{2M}\sum_{t,n}m_{t,n}\left\|D_{p}^{-1}(\widehat{\mathbf{p}}_{t,n}-\mathbf{p}_{t,n})\right\|_{1}.
\label{eq:position_loss}
\end{equation}
\begin{equation}
\ell_{\mathrm{vel}}=\frac{1}{2M}\sum_{t,n}m_{t,n}\left\|D_{v}^{-1}(\widehat{\mathbf{v}}_{t,n}-\mathbf{v}_{t,n})\right\|_{1}.
\label{eq:velocity_loss}
\end{equation}
The sums run over future frames and agent slots.
The factor of two averages over spatial coordinates, and validity masking excludes padded or invalid reference states.

The endpoint loss supervises the final available position of each agent.
Let $\mathcal{V}=\{n:\sum_{t}m_{t,n}>0\}$ and $t_{n}^{\star}=\max\{t:m_{t,n}=1\}$ for $n\in\mathcal{V}$.
For $|\mathcal{V}|>0$, we use
\begin{equation}
\ell_{\mathrm{end}}=\frac{1}{2|\mathcal{V}|}\sum_{n\in\mathcal{V}}\left\|D_{p}^{-1}(\widehat{\mathbf{p}}_{t_{n}^{\star},n}-\mathbf{p}_{t_{n}^{\star},n})\right\|_{1}.
\label{eq:endpoint_loss}
\end{equation}
This target remains defined when an agent becomes invalid before the rollout ends.
A class balanced binary cross entropy loss $\ell_{\mathrm{valid}}$ supervises the validity logits using every future slot, including padded slots.
The dynamics objective is
\begin{equation}
\ell_{\mathrm{resp}}=\ell_{\mathrm{pos}}+0.5\ell_{\mathrm{vel}}+\ell_{\mathrm{end}}+0.1\ell_{\mathrm{valid}}.
\label{eq:response_loss}
\end{equation}
The four terms supervise the spatial trajectory, physical velocity, final valid position, and state validity, respectively.

\subsection{Joint Optimization}
\label{app:optimization}

\myparagraph{Objective.}For each training example and sampled flow time, the video loss is
\begin{equation}
\ell_{\mathrm{video}}=\frac{1}{D}\left\|\mathbf{u}_{\theta}(\mathbf{z}_{\tau},\tau;\mathbf{z}_{\mathrm{obs}},\mathbf{A},\mathbf{C})-(\boldsymbol{\epsilon}-\mathbf{z})\right\|_{2}^{2}.
\label{eq:app_video_loss}
\end{equation}
The conditions are computed as $\mathbf{C}=G_{\psi}(F_{\phi}(\mathbf{z}_{\mathrm{obs}},\mathbf{A}))$.
Here, $\mathbf{z}_{\tau}=(1-\tau)\mathbf{z}+\tau\boldsymbol{\epsilon}$, and $D$ is the number of scalar entries in the future video latent tensor.
The norm covers only these future entries.
Observed video, ego action, and response tokens receive no flow target.
The complete objective averages over training examples, Gaussian noise, and the backbone flow schedule:
\begin{equation}
\mathcal{L}=\mathbb{E}[\ell_{\mathrm{video}}+\lambda_{\mathrm{resp}}\ell_{\mathrm{resp}}], \qquad \lambda_{\mathrm{resp}}=1.
\label{eq:joint_loss}
\end{equation}

\myparagraph{Gradient coupling.}The video objective updates the dynamics module through the estimated agent states and the response encoder.
For vectorized states and tokens, define $J_{F}=\partial\widehat{\mathbf{S}}/\partial\phi$ and $J_{G}=\partial\mathbf{C}/\partial\widehat{\mathbf{S}}$.
For a sample objective $\ell_{\mathrm{joint}}=\ell_{\mathrm{video}}+\lambda_{\mathrm{resp}}\ell_{\mathrm{resp}}$, the chain rule gives
\begin{equation}
\nabla_{\phi}\ell_{\mathrm{joint}}=J_{F}^{\top}\!\left(J_{G}^{\top}\nabla_{\mathbf{C}}\ell_{\mathrm{video}}+\lambda_{\mathrm{resp}}\nabla_{\widehat{\mathbf{S}}}\ell_{\mathrm{resp}}\right).
\label{eq:app_gradient_coupling}
\end{equation}
The expression holds at differentiable points, with subgradients used for the absolute error terms where needed.
It identifies the two sources of supervision acting on the same agent state estimates.
The dynamics objective measures agreement with reference agent states, while the video objective trains these estimates to support visual generation.
Both the regression channels and the validity probabilities participate in this differentiable conditioning pathway.

\myparagraph{Initialization and warmup.}We initialize the response conditioning pathway from a checkpoint trained with reference future tracks.
This checkpoint provides a learned mapping from structured agent states to visual generation.
We then freeze the video generator and optimize the internal dynamics module for 500 iterations using $\ell_{\mathrm{resp}}$.
During warmup, the video generator continues to receive conditions constructed from reference tracks.

\myparagraph{Joint training.}After warmup, we replace reference conditions with response tokens computed from the model's own state estimates in Equation~\eqref{eq:app_video_loss} and optimize the joint objective for 6000 iterations.
This stage updates the dynamics module, response projection, response modality embedding, and rank-16 LoRA adapters~\citep{hu2021lora} on the video transformer's attention projections.
The remaining Cosmos3 parameters and the video tokenizer are frozen.
Agent state estimates remain connected to the video objective through the response encoder.
Other backbone training and sampling settings follow the Cosmos3 configuration~\citep{agarwal2026cosmos}.
We train the two ablation variants separately, removing the reactive dynamics prior in one and position information in the other.

\subsection{Simulation}
\label{app:inference}

The tokenizer first encodes the observed video.
\reactdrive{} then computes physical agent state estimates and maps them to response tokens:
\begin{equation}
\widehat{\mathbf{S}}=F_{\phi}(\mathbf{z}_{\mathrm{obs}},\mathbf{A}), \qquad \mathbf{C}=G_{\psi}(\widehat{\mathbf{S}}).
\label{eq:reactdrive_inference}
\end{equation}
Both quantities are computed once for each observation and ego query.
The response tokens remain fixed during video sampling. The stochastic video rollout starts from $\widehat{\mathbf{z}}_{1}\sim\mathcal{N}(\mathbf{0},\mathbf{I})$ and follows
\begin{equation}
\frac{d\widehat{\mathbf{z}}_{\tau}}{d\tau}=\mathbf{u}_{\theta}(\widehat{\mathbf{z}}_{\tau},\tau;\mathbf{z}_{\mathrm{obs}},\mathbf{A},\mathbf{C}).
\label{eq:app_flow_sampling}
\end{equation}
We integrate from $\tau=1$ to $\tau=0$ while holding the observations, ego actions, and cached response tokens fixed.
The frozen decoder reconstructs $[\mathbf{z}_{\mathrm{obs}};\widehat{\mathbf{z}}_{0}]$, and we retain the $T$ future frames as $\widehat{\mathbf{Y}}$.
All response tokens are computed from observed history and the specified ego trajectory.

\subsection{Kinematic Interpretation of Dynamics Supervision}
\label{app:kinematic_grounding}

We relate position and velocity errors to deviations from the reference trajectories' local kinematic relations.
Consider adjacent states of the same agent with $m_{t,n}=m_{t+1,n}=1$, separated by $\Delta t$ in the fixed reference frame.
Define the reference kinematic residual as
\begin{equation}
\mathbf{r}_{t,n}=\mathbf{p}_{t+1,n}-\mathbf{p}_{t,n}-\Delta t\,\mathbf{v}_{t,n}.
\label{eq:reference_kinematic_residual}
\end{equation}
The estimated residual $\widehat{\mathbf{r}}_{t,n}$ is defined by replacing the reference positions and velocities with their estimated values.
These residuals measure departure from a local constant-velocity approximation; the reference residual need not be zero.

\myparagraph{Kinematic error bound.}Let $\mathbf{e}^{p}_{t,n}=\widehat{\mathbf{p}}_{t,n}-\mathbf{p}_{t,n}$ and $\mathbf{e}^{v}_{t,n}=\widehat{\mathbf{v}}_{t,n}-\mathbf{v}_{t,n}$.
Define their normalized magnitudes as $\varepsilon^{p}_{t,n}=\|D_p^{-1}\mathbf{e}^{p}_{t,n}\|_1$ and $\varepsilon^{v}_{t,n}=\|D_v^{-1}\mathbf{e}^{v}_{t,n}\|_1$.
Then
\begin{equation}
\left\|D_p^{-1}(\widehat{\mathbf{r}}_{t,n}-\mathbf{r}_{t,n})\right\|_1\leq\varepsilon^{p}_{t+1,n}+\varepsilon^{p}_{t,n}+\Delta t\,\kappa\varepsilon^{v}_{t,n}.
\label{eq:kinematic_error_bound}
\end{equation}
Here, $\kappa=\|D_p^{-1}D_v\|_{1\rightarrow1}$.
Our normalization uses $D_p=\operatorname{diag}(100,50)$ and $D_v=\operatorname{diag}(30,30)$, giving $\kappa=0.6$.

\myparagraph{Proof.}Subtracting the two residuals gives
\begin{equation}
\widehat{\mathbf{r}}_{t,n}-\mathbf{r}_{t,n}=\mathbf{e}^{p}_{t+1,n}-\mathbf{e}^{p}_{t,n}-\Delta t\,\mathbf{e}^{v}_{t,n}.
\label{eq:kinematic_residual_difference}
\end{equation}
Equation~\eqref{eq:kinematic_error_bound} follows from the triangle inequality and the induced matrix norm.

The reference residual also has a physical interpretation.
If the reference states sample a twice-differentiable trajectory with $\dot{\mathbf{p}}=\mathbf{v}$ and $\|\dot{\mathbf{v}}(\xi)\|_1\leq a_{\max}$ over the interval, integration yields
\begin{equation}
\|\mathbf{r}_{t,n}\|_1\leq\frac{1}{2}a_{\max}\Delta t^2.
\label{eq:reference_acceleration_bound}
\end{equation}
Thus, position and velocity supervision controls deviation from the reference trajectory's local kinematic relation.
In our 10\,Hz setting, $\Delta t=0.1$\,s.

This analysis applies to estimated physical agent states at adjacent frames with valid reference supervision.
It introduces no additional residual loss and does not establish the physical correctness of generated observations or responses to unobserved ego actions.

\section{Evaluation Metrics}
\label{app:evaluation_metrics}

This section defines the metrics used in Section~\ref{sec:experimental_setup}.
We use the same feature extractors, trajectory-recovery procedure, detector, matching strategy, and aggregation rules for all evaluated methods.

\subsection{Visual Quality}
\label{app:appearance_metrics}

Let $\{\mathbf{z}^{r}_{n}\}_{n=1}^{N_r}$ and $\{\mathbf{z}^{g}_{n}\}_{n=1}^{N_g}$ denote features extracted from real and generated samples, respectively.
We compute their empirical means and covariance matrices as
\begin{equation}
\boldsymbol{\mu}_{r}
=
\frac{1}{N_r}
\sum_{n=1}^{N_r}
\mathbf{z}^{r}_{n},
\qquad
\boldsymbol{\mu}_{g}
=
\frac{1}{N_g}
\sum_{n=1}^{N_g}
\mathbf{z}^{g}_{n},
\end{equation}
and
\begin{equation}
\boldsymbol{\Sigma}_{r}
=
\frac{1}{N_r-1}
\sum_{n=1}^{N_r}
\left(
\mathbf{z}^{r}_{n}-\boldsymbol{\mu}_{r}
\right)
\left(
\mathbf{z}^{r}_{n}-\boldsymbol{\mu}_{r}
\right)^{\top},
\end{equation}
\begin{equation}
\boldsymbol{\Sigma}_{g}
=
\frac{1}{N_g-1}
\sum_{n=1}^{N_g}
\left(
\mathbf{z}^{g}_{n}-\boldsymbol{\mu}_{g}
\right)
\left(
\mathbf{z}^{g}_{n}-\boldsymbol{\mu}_{g}
\right)^{\top}.
\end{equation}

The Fr\'{e}chet distance between the two feature distributions is
\begin{equation}
\mathcal{D}_{\mathrm{Fr\acute{e}chet}}
=
\left\|
\boldsymbol{\mu}_{r}
-
\boldsymbol{\mu}_{g}
\right\|_{2}^{2}
+
\operatorname{Tr}
\left(
\boldsymbol{\Sigma}_{r}
+
\boldsymbol{\Sigma}_{g}
-
2
\left(
\boldsymbol{\Sigma}_{r}
\boldsymbol{\Sigma}_{g}
\right)^{1/2}
\right).
\label{eq:frechet_distance}
\end{equation}

FID~\citep{heusel2017gans} applies Eq.~\eqref{eq:frechet_distance} to image features extracted from individual frames.
FVD~\citep{unterthiner2018towards} applies the same distance to spatiotemporal features extracted from complete video clips.
Both metrics compare generated and recorded videos under recorded ego trajectories.

\subsection{Ego Controllability}
\label{app:ego_metrics}

For each generated video, we use VGGT-$\Omega$~\citep{wang2026vggtomega} to estimate camera poses and recover the ego trajectory.
We compare the recovered trajectory with the specified ego trajectory~\citep{arai2024act,zhou2026drivinggen}. Ego ADE averages the displacement error over future frames, while Ego FDE measures the error at the final frame.

\subsection{Agent Responses under Recorded Ego Trajectories}
\label{app:agent_metrics}

Under recorded ego trajectories, Agent ADE/FDE compare matched vehicle tracks in the generated and recorded videos.
Vehicles are detected using YOLO26x~\citep{jocher2026ultralytics}, and tracks are associated using Hungarian matching~\citep{kuhn1955hungarian}.
We measure bounding-box center displacement in image space, normalized by the corresponding box height at the anchor frame.
For each matched track, ADE averages the displacement error over frames with detections in both videos. FDE uses the last such frame.
The reported scores average these errors across matched tracks.
Section~\ref{sec:human_evaluation} evaluates agent responses to edited ego actions through human judgments.

\subsection{Collision Rate}
\label{app:collision_rate}

Under edited ego trajectories, we use YOLO26~\citep{jocher2026ultralytics} to detect vehicles in sampled future frames.
A frame is flagged for a potential ego collision if a detected vehicle's bounding box overlaps the region immediately ahead of the ego vehicle in the image.
CR is the percentage of sampled future frames flagged by this rule. We report CR overall and for each maneuver group.

\section{NAVSIM Planning Evaluation}
\label{app:navsim_evaluation}

This section describes the data augmentation and planner training settings used in Section~\ref{sec:planning}.
We follow the NAVSIM protocol~\citep{dauner2024navsim} for trajectory targets and planning evaluation, using the front camera as the visual input.

\subsection{Training Data and Augmentation}
\label{app:navsim_splits}

We use the official NAVTRAIN and NAVTEST splits for planner training and evaluation, respectively.
NAVTEST is excluded from augmentation, model selection, and checkpoint selection.
The three training settings use real data alone, real data augmented with Cosmos3 observations, and real data augmented with \reactdrive{} observations.

Each generator contributes 500 videos constructed from the same source scenes and edited ego trajectories.
The generated observations supplement the real training data.
Both augmentation settings use the same procedure to construct planner inputs and trajectory labels, with the generation method as the only difference between them.

\subsection{Observations and Trajectory Targets}
\label{app:navsim_sensor_replacement}

The front camera provides the visual input to all three planner variants.
For augmented samples, this input is obtained from the corresponding generated video.

Following NAVSIM~\citep{dauner2024navsim}, each planning sample is anchored at its observation time $t$.
The target contains eight future ego poses sampled at $0.5$-second intervals over a $4$-second horizon, expressed in the ego coordinate system at time $t$.
For real samples, these targets are obtained from the recorded ego trajectory.
For augmented samples, they are constructed from the ego trajectory supplied to the generator using the same temporal and coordinate conventions.
Specifically, an observation at time $t$ is paired with the subsequent ego poses at $t+0.5,\ldots,t+4.0$ seconds.
Corresponding samples from Cosmos3 and \reactdrive{} share the same trajectory targets.

\subsection{Planner Training}
\label{app:navsim_training}

We train a separate TransFuser~\citep{chitta2022transfuser} planner for each training setting.
All variants share the same initialization, optimizer, learning-rate schedule, input configuration, and checkpoint-selection rule.
The two augmented variants additionally share the source scenes, conditioning ego trajectories, and sample-construction procedure.
The trained planners are evaluated on NAVTEST using the official NAVSIM evaluator.

\subsection{NAVSIM Metrics}
\label{app:navsim_metrics}

We use the official NAVSIM evaluator~\citep{dauner2024navsim} to assess collision avoidance, road compliance, driving efficiency, and comfort.
No-at-fault collisions (NC) scores the avoidance of collisions attributed to the ego vehicle, while time-to-collision compliance (TTC) assesses whether the simulated trajectory satisfies the prescribed collision-time bound.
Drivable-area compliance (DAC) measures whether the vehicle remains within the drivable area, and driving-direction compliance (DDC) evaluates whether it follows the permitted driving direction.
Ego progress (EP) measures normalized distance traveled along the route.
Dynamic comfort (C) evaluates acceleration, jerk, and turning dynamics against the benchmark's comfort limits.

These scores are normalized to $[0,1]$ for each scene.
The Predictive Driver Model Score (PDMS) combines NC and DAC as multiplicative penalties with a weighted average of TTC, EP, and comfort:
\begin{equation}
\mathrm{PDMS}_{i}=\mathrm{NC}_{i}\,\mathrm{DAC}_{i}\,\frac{5\,\mathrm{TTC}_{i}+5\,\mathrm{EP}_{i}+2\,\mathrm{C}_{i}}{12}.
\label{eq:navsim_pdms}
\end{equation}
We use the default NAVSIM weights and report DDC separately.
PDMS is computed for each scene before averaging across NAVTEST.
All component scores are likewise averaged across scenes, and all reported scores are scaled by $100$.

\section{Human Evaluation}
\label{app:human_evaluation}

\subsection{Study Design}

The human evaluation received ethics approval.
The study includes 34 raters and 50 evaluation cases.
For each case, the video pool contains one \reactdrive{} video and one video from each of the five baselines: Vista, GEM, Epona, DrivingWorld, and Cosmos3.
The pool therefore contains 300 videos, with 50 videos per model.
Each comparison pairs \reactdrive{} with one baseline for the same case.
Each rater is assigned 20 randomly sampled comparisons.
Model identities are hidden throughout the evaluation, and the left and right positions of the videos are randomized.
Of the 680 assigned comparisons, participants completed 623. The remaining 57 belong to unfinished sessions.
Each completed comparison provides one judgment for each of the four criteria: agent reactivity, ego controllability, visual quality, and overall preference.

\subsection{Criteria and Aggregation}

Agent reactivity concerns whether surrounding agents respond appropriately to the ego action.
Ego controllability concerns whether the generated ego trajectory follows the requested action.
Visual quality concerns the plausibility of the generated video.
Overall preference records which video the rater prefers when considering the comparison as a whole.
Table~\ref{tab:human_evaluation} reports 623 completed judgments per criterion.
We pool these judgments across the five baselines and report the fraction favoring \reactdrive{}, giving each completed comparison equal weight.

\section{Visualization Results}
\label{app:visualization}

Additional video examples and comparisons are provided in the supplementary webpage.
To view them, extract the supplementary ZIP file and open \texttt{index.html} in the webpage folder using a web browser, keeping all accompanying files and folders in their original locations.
The video gallery supports filtering by dataset and ego action.

\end{document}